\documentclass[journal]{IEEEtran}
\usepackage{amsmath}    
\usepackage{graphicx}   
\usepackage{color}      
\usepackage{hyperref}
\usepackage{comment}
\usepackage{subfigure}
\usepackage{times}
\usepackage{psfig}
\usepackage{epsfig}
\usepackage{latexsym}
\usepackage{bbold,bbm}
\usepackage{setspace}
\usepackage{amssymb,amsmath}
\usepackage{mathrsfs}
\usepackage{amsgen,amsfonts,amsbsy,amsthm}
\usepackage{algorithmic,algorithm}
\usepackage{multirow} 
\usepackage{array}
\usepackage{enumitem}
\usepackage{bbding}
\usepackage{pifont}
\usepackage{wasysym}
\usepackage{amssymb}

\usepackage{amsfonts}
\usepackage{amssymb}
\usepackage{multirow}
\usepackage{times}
\usepackage{epsfig}
\usepackage{latexsym}
\usepackage{bbold,bbm,dsfont}
\usepackage{setspace}
\usepackage{amssymb,amsmath}
\usepackage{mathrsfs}
\usepackage{amsgen,amsfonts,amsbsy,amsthm}
\usepackage{algorithmic,algorithm}
\usepackage{cite}
\usepackage{calc}
\usepackage{color}
\usepackage{gensymb} 

\usepackage{hyperref}
\usepackage{times}
\usepackage{subfigure}
\usepackage{longtable}
\DeclareGraphicsExtensions{.eps}
\usepackage[all,2cell]{xy} \UseAllTwocells \SilentMatrices
\newcommand{\no}{\nonumber}
\newcommand{\beq}{\begin{equation}}
\newcommand{\eeq}{\end{equation}}

\newcommand{\be}{\begin{equation}}
\newcommand{\ee}{\end{equation}}
\newcommand{\bea}{\begin{eqnarray}}
\newcommand{\eea}{\end{eqnarray}}
\newcommand{\bean}{\begin{eqnarray*}}
\newcommand{\eean}{\end{eqnarray*}}

\newcommand{\bit}{\begin{itemize}}
\newcommand{\eit}{\end{itemize}}
\newcommand{\ben}{\begin{enumerate}}
\newcommand{\een}{\end{enumerate}}

\renewcommand\footnotemark{}

\definecolor{dark-green}{rgb}{0.0, 0.5, 0.0}

\begin{document}
\title{Animation and Chirplet-Based Development of a PIR Sensor Array for Intruder Classification in an Outdoor Environment } %

%
%
%



\author{
\authorblockN{ Raviteja Upadrashta, Tarun Choubisa, A. Praneeth, Tony G., V. S. Aswath, P. Vijay Kumar, Fellow, IEEE}
\authorblockA{Dept.\ of Electrical  Comm. Engg., Indian Institute of Science, Bengaluru, 560012, India}\\
\authorblockN{Sripad Kowshik, Hari Prasad Gokul R, T. V. Prabhakar}\\
\authorblockA{Dept.\ of Electronic Systems Engg., Indian Institute of Science, Bengaluru, 560012, India}
}

\maketitle
\begin{abstract}
This paper presents the development of a passive infra-red sensor tower platform along with a classification algorithm to distinguish between human intrusion, animal intrusion and clutter arising from wind-blown vegetative movement in an outdoor environment. The research was aimed at exploring the potential use of wireless sensor networks as an early-warning system to help mitigate human-wildlife conflicts occurring at the edge of a forest. There are three important features to the development.  Firstly, the sensor platform employs multiple sensors arranged in the form of a two-dimensional array to give it a key spatial-resolution capability that aids in classification.  Secondly, given the challenges of collecting data involving animal intrusion, an Animation-based Simulation tool for Passive Infra-Red sEnsor (ASPIRE) was developed that simulates signals corresponding to human and animal intrusion and some limited models of vegetative clutter.  This speeded up the process of algorithm development by allowing us to test different hypotheses in a time-efficient manner.  Finally, a chirplet-based model for intruder signal was developed that significantly helped boost classification accuracy despite drawing data from a smaller number of sensors.  An SVM-based classifier was used which made use of chirplet, energy and signal cross-correlation-based features. The average accuracy obtained for intruder detection and classification on real-world and simulated data sets was in excess of $97\%$.
\end{abstract}
\vspace*{-0.25in}


\section{Introduction} \label{sec:introduction} 
As described above, this project\footnote{The work presented here was supported in part by an Indo-US project jointly funded by the US National Science Foundation and the Indian Department of Electronics and Information Technology.} concerns the development of a Passive Infra-Red (PIR) based Sensor Tower Platform (STP) that can distinguish between humans, animals and wind-blown vegetative clutter in an outdoor setting. We limit our attention to a small subclass of animals that are comparable in size and shape to a tiger or a dog. In the sequel\footnote{An earlier version of this paper was presented at the ISSNIP-2015 Conference. DOI:10.1109/ISSNIP.2015.7106914}, we will refer to this sub-class of animals simply as animals. Also, when we speak of an intruder, the intruder could be either human or animal. We will refer to clutter generated by wind-blown vegetation as simply clutter. Power is not commonly available in settings of the type considered here and this motivates the use of low-power sensor networks.  
  \vspace*{-0.15in}
\subsection{Prior Work}
PIR sensors for detecting human motion in outdoor environments has only recently been investigated \cite{arora2005exscal,  gu2005lightweight, SmartDetect, hong2013reduction, jacobs2009pyroelectric}.  Hong et. al. in \cite{hong2013reduction} use digital PIR sensors and energy thresholding for detecting human motion. The amount of signal processing that can be done using digital PIR sensors is limited in comparison to analog sensors that allow more sophisticated signal-processing algorithms to be employed for rejecting clutter. In \cite{gu2005lightweight}, the PIR signal is first high-pass filtered to remove low frequency components resulting from slow environment changes and the signal energy is then compared against an adaptive threshold. In \cite{SmartDetect}, clutter signals arising from environmental changes and wind-blown vegetation are rejected by a Support Vector Machine (SVM) based classifier that uses a Haar transform-based feature vector.  However, the articles \cite{arora2005exscal,  SmartDetect, hong2013reduction, gu2005lightweight} described above, do not consider the problem of distinguishing between human and animal intrusions. 

The authors of \cite{jacobs2009pyroelectric} develop a Sensor Platform (SP) that is capable of classifying between human and animal that makes use of relatively expensive germanium lenses along with high resolution PIR sensors.  The focus of the work presented here is on the development of a STP that makes use of off-the-shelf, relatively inexpensive PIR sensors and lenses. In a publication that appeared after the appearance of our 2015 conference publication \cite{upadrashta2015animation}, Zhao et. al. in \cite{zhao2016emd} investigated the problem of discriminating between humans and false alarms generated by motion of animals of shape similar to dogs and geese using a single PIR sensor in conjunction with a multi-lens. However, false alarms potentially arising from moving vegetative clutter are not treated in this paper. 

Section~\ref{sec:background} provides background on the PIR sensor. Section~\ref{sec:design_approach} gives a detailed description of our approach to PIR STP design.  Section~\ref{sec:data_collection} describes our data collection efforts. Section~\ref{sec:animation}, provides a description of ASPIRE, an animation-based simulation tool that was used to simulate PIR signals generated by human and animal intrusions.   Chirp-based modeling of the intruder signal and its use in classification is covered in Section~\ref{sec:chirp}.  Other features employed for detection and classification are described in Section~\ref{sec:other_features}.   Our classification algorithm and experimental results are presented in Section~\ref{sec:results}.  The final section draw conclusions and presents thoughts on future extensions of this work. 
\section{PIR Sensor Background} \label{sec:background}

PIR sensors work on the principle of pyroelectricity which causes them to respond to changes in incident radiation  \cite{lang2005pyroelectricity}. This ability has been used in a wide range of applications such as temperature sensing, traffic control, fire alarms, thermal imaging, and radiometers \cite{whatmore1986pyroelectric}.  A further application involves the detection of humans as there are estimates that place the heat radiated by a human up to as much as 100 W.  The use of PIR sensors in detecting human motion has been restricted mainly to indoor applications ranging from home security, smart homes, health care, hallway monitoring, gesture recognition, walker recognition, etc.  As noted below in Section~\ref{sec:challenges}, there are challenges to be overcome when attempting to employ a PIR in an outdoor environment.

\subsection{Block Diagram of a PIR Sensor}
\begin{figure}[h]
    \centering
    \vspace*{-0.05in} 
	\includegraphics[trim= 0in  0.5in 0in 0in, width=2in]{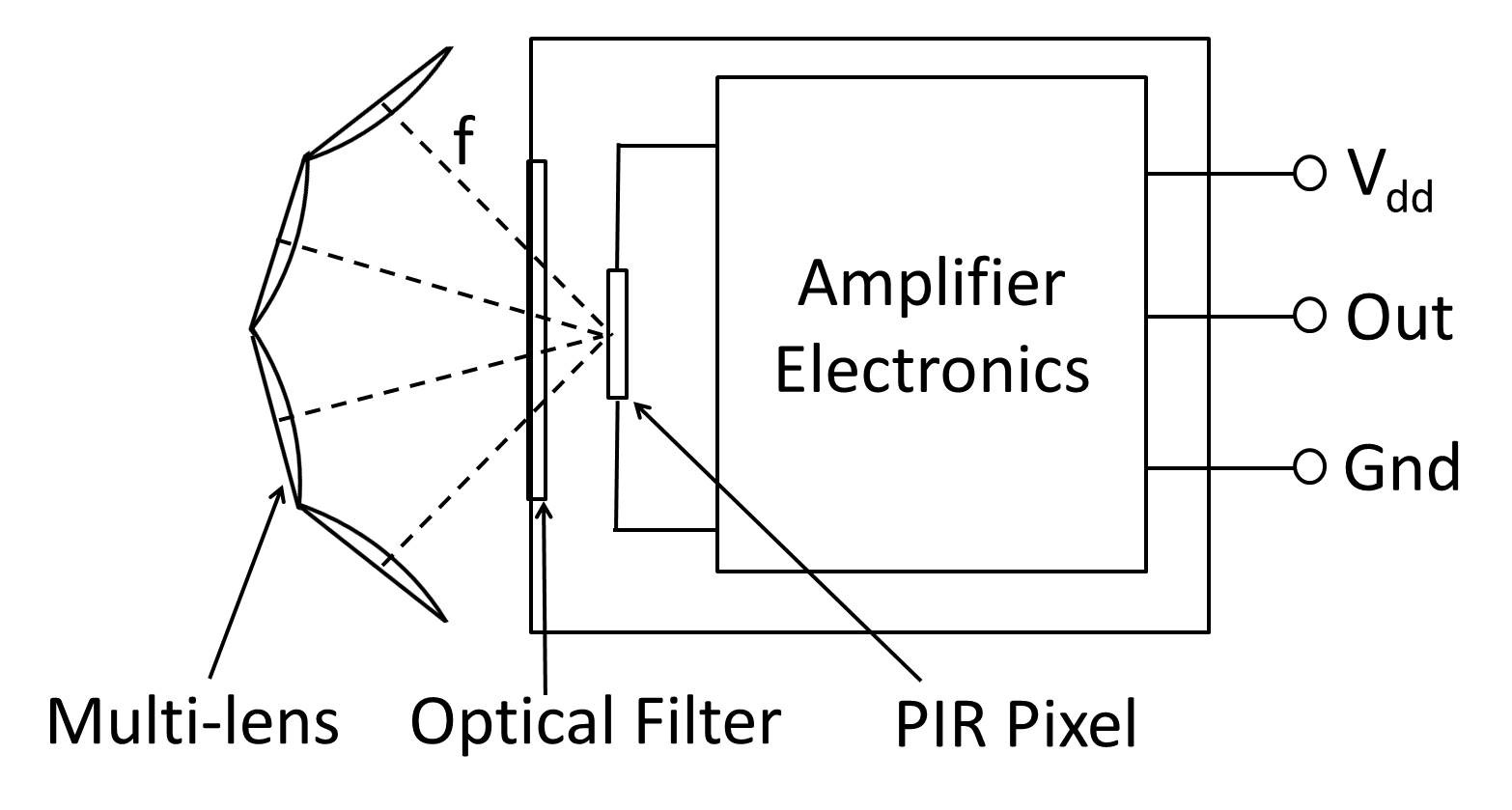} 
	\vspace*{-0.05in} 
    \caption{The block diagram of a typical PIR sensor. \label{fig:PIRBlockDiagram}}
\end{figure}

\begin{figure}[h]
  \centering
  \subfigure[]{\label{fig:VPA}\includegraphics[trim= 0in  0in 0in 0in, height=0.9in]{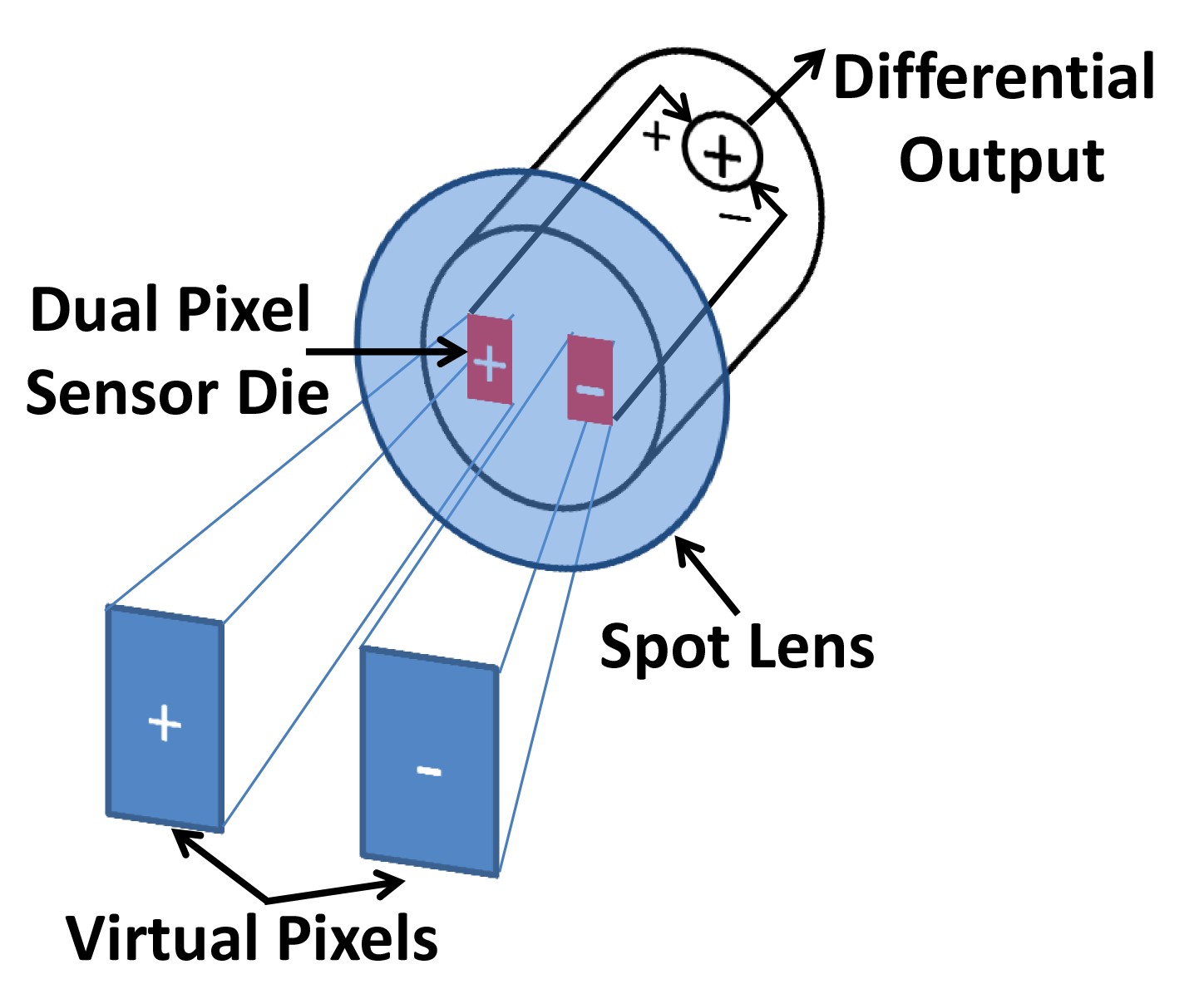}} 
    \hspace{0.5cm}
    \subfigure[]{\label{fig:signal}\includegraphics[trim= 0in  0in 0in 0in, height=0.9in]{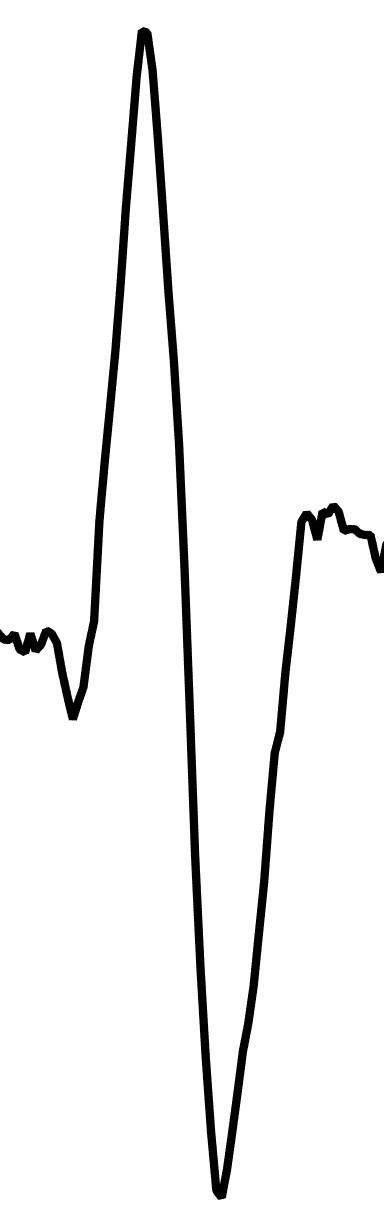}}
    \subfigure[]{\label{fig:VPA_multilens}\includegraphics[trim= 0in  0.5in 0in 0in, width=2in]{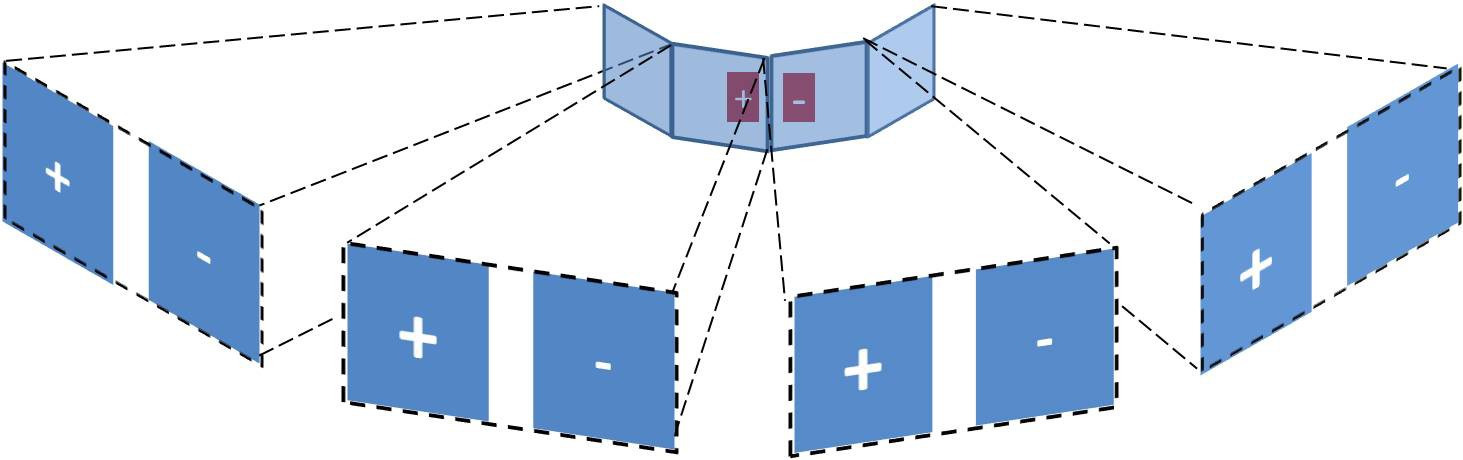}}
    \subfigure[]{\label{fig:signal_multilens}\includegraphics[trim= 0in  0.4in 0in 0in, width=1.5in]{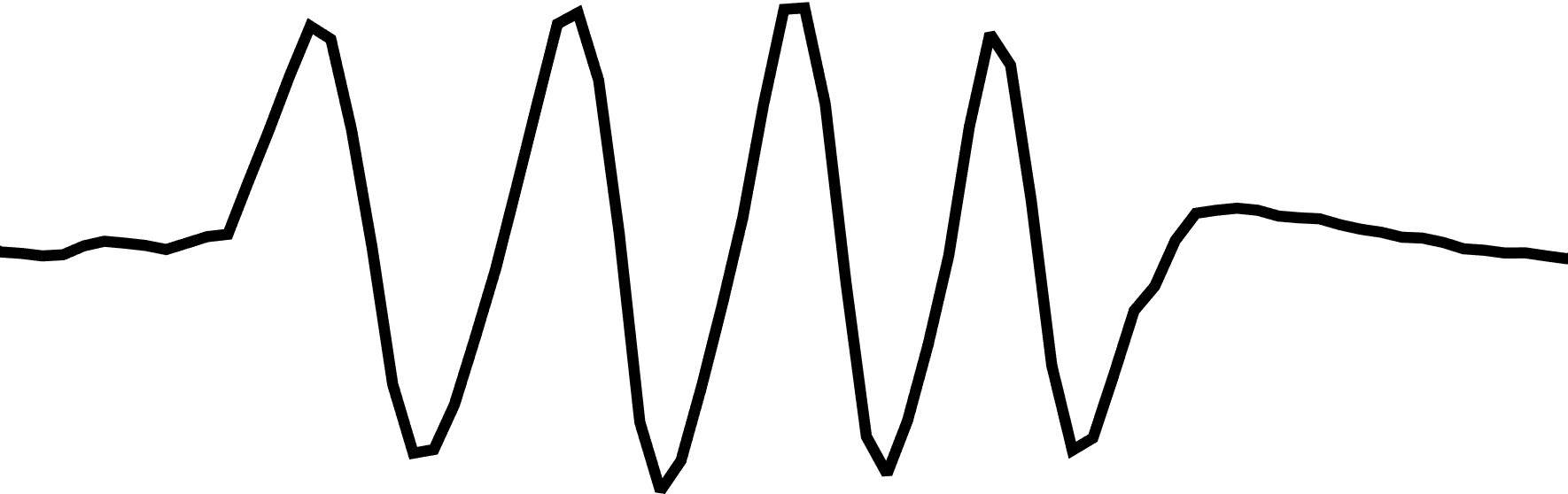}}
    \caption{The VPA associated with a $(1 \times 2)$-pixel array and a spot lens is shown in (a).   The corresponding signal generated when an intruder moves in a plane parallel to the face of the sensor is shown in (b).  When the spot lens is replaced by a multi-lens, we obtain the VPA shown in (c) and the signal waveform displayed in (d). }
\end{figure}

The block diagram of a typical, commercially-available PIR sensor is shown in Fig.~\ref{fig:PIRBlockDiagram}. 
The PIR sensor typically consists of several radiation-sensing elements, called pixels, arranged in the form of an array, a common form being a $(1 \times 2)$ arrangement as shown in Fig.~\ref{fig:VPA}. We will refer to this $(1 \times 2)$ PIR sensor, simply as a PIR sensor.  The pair of pixels in the $(1 \times 2)$ arrangement are wired together in a differential manner to overcome false alarms triggered by changes in the ambient temperature. A lens system is typically used to expand the Field of View (FoV) of the sensor and plays a key role in defining the sensor output signal. The lens system serves to focus radiation from different directions onto the pixels and consists of either a single lens or a multi-lens.  A multi-lens is a set of contiguous lenses sharing a common focal point.  The multi-lens shown in Fig.~\ref{fig:PIRBlockDiagram} corresponds to a set of contiguous plano-convex lenses~\footnote{In practice, each plano-convex lens is replaced by a Fresnel lens to reduce attenuation of the incident radiation.}.   The FoV of the sensor can be viewed as a set of virtual beams cast out into space along which radiation is received. Fig.~\ref{fig:VPA} illustrates the virtual beams cast out by a single spot lens, when placed in front of a $(1 \times 2)$ sensor.   Given any plane in 3-dimensional space, the intersection of the FoV of the sensor with the plane is called the Virtual Pixel Array (VPA) associated with the particular plane.  Fig.~\ref{fig:VPA} shows the VPA associated to the $(1 \times 2)$ sensor used in conjunction with a spot lens. A typical signal generated as a result of a human intruder moving across the FoV of  $(1 \times 2)$ PIR pixel array is shown in Fig. ~\ref{fig:signal}.  When the spot lens is replaced by a multi-lens, the resulting virtual beams are shown in Fig.~\ref{fig:VPA_multilens}. The output of the sensor in this case is the superposition of the signals generated by each pair of virtual pixels. A typical signal output for such an arrangement is shown in Fig.~\ref{fig:signal_multilens}. 


The human being radiates the peak emission at a wavelength of roughly 10 $\mu$m.  An optical filter is used to reduce the influence of moving objects radiating at wavelengths significantly different from those of a human being. 
\subsection{PIR Signal Generation}  \label{sec:PIR_sensor} 
\vspace*{-0.12in}
\subsubsection{Radiation Model} \label{sec:RadiationModel}
The first step in determining the signal generated by a PIR sensor is to estimate the net radiation received from the source. Assuming the source can be modelled as a Lambertian source, i.e., a source that radiates equal energy in all directions, the net power transfer is \cite{RadiationModel}
\beq w(t) = \frac{\tau \eta F A_{\text{e}} A_{\text{proj}}(t)\sigma (T_{\text{obj}}^4-T_{\text{b}}^4)}{\pi R^2},
\label{eq:radiation} 
\eeq
where, $\tau$ captures attenuation due to the atmosphere, $\eta$ is the transmission coefficient of the lens, $F$ is the fraction of the total radiation within the bandwidth of the optical filter, $A_{\text{e}}$ is the area of the lens aperture, $A_{\text{proj}}(t)$ is the area of the source when projected on to the VPA, $\sigma$ is the Stefan-Boltzmann's constant, $T_{\text{obj}}$ is the object temperature in Kelvin, $T_{\text{b}}$ is the background temperature in Kelvin and $R$ is the distance of the radiating source from the detector.  

\subsubsection{Impulse Response}

One can model a PIR sensor as a bandpass filter with an impulse response given by \cite{hossain1991pyroelectric, arpan2012WCSN}
\beq
h(t)=k_1e^{-k_2t}-k_3e^{-k_4t},
\eeq
where, the constants $k_1$ through $k_4$ depend on the physical properties of the PIR crystal and the amplifier electronics. The output voltage $v(t)$ of the PIR sensor in response to a time-varying incident input radiation $w(t)$ can be computed by simply convolving $w(t)$ with $h(t)$ i.e., $v(t)=w(t)\ast h(t)$.

%
%

 Equation~\eqref{eq:radiation} and the above model of a PIR sensor are employed by ASPIRE to simulate the output signal of a PIR sensor as will be seen in Section~\ref{sec:animation}. Simulation of the signal is useful when we wish to quickly understand the efficacy of a particular lens design without expending the time and effort needed for real-world data collection.
  \vspace*{-0.1in}
\subsection{Challenges Faced in an Outdoor Deployment of a PIR Sensor} \label{sec:challenges} 
One of the challenges to be overcome in an outdoor setting is the need to detect and classify in the presence of clutter. In such a setting, the radiation incident on a PIR could be altered due to changes in the environment, for example, when leaves blow in the wind or the sun comes out from behind a cloud. 

  \vspace*{-0.05in}
\section{Approach to PIR Sensor Platform Design} \label{sec:design_approach} 

Our two principal objectives are rejecting false alarms arising from moving vegetation and distinguishing between human and animal intruder motion. Two simple observations were key to the development of the STP.   Firstly, that it is possible to distinguish between human and animal based on their height.  Secondly, intruder motion is translational in nature, while vegetative motion tends to be oscillatory.   

The following assumptions were made: 
\ben
\item At any given time, only a single intruder will be present within the FoV of the sensor; 
\item Intruders move in a straight line at a uniform speed that is in a specified range: $1$ to $3$ m/sec;
\item Only animals having the same physical features as either a dog, wolf, leopard or tiger are considered here. 
\een 

\subsection{VPA Design}
\begin{figure}[h]
\centering
  \vspace*{-0.2in}
\subfigure[Sensor platform design]{\label{fig:Final_Design}\includegraphics[trim= 0.3in  0.75in 0.15in 0in, height=1in]{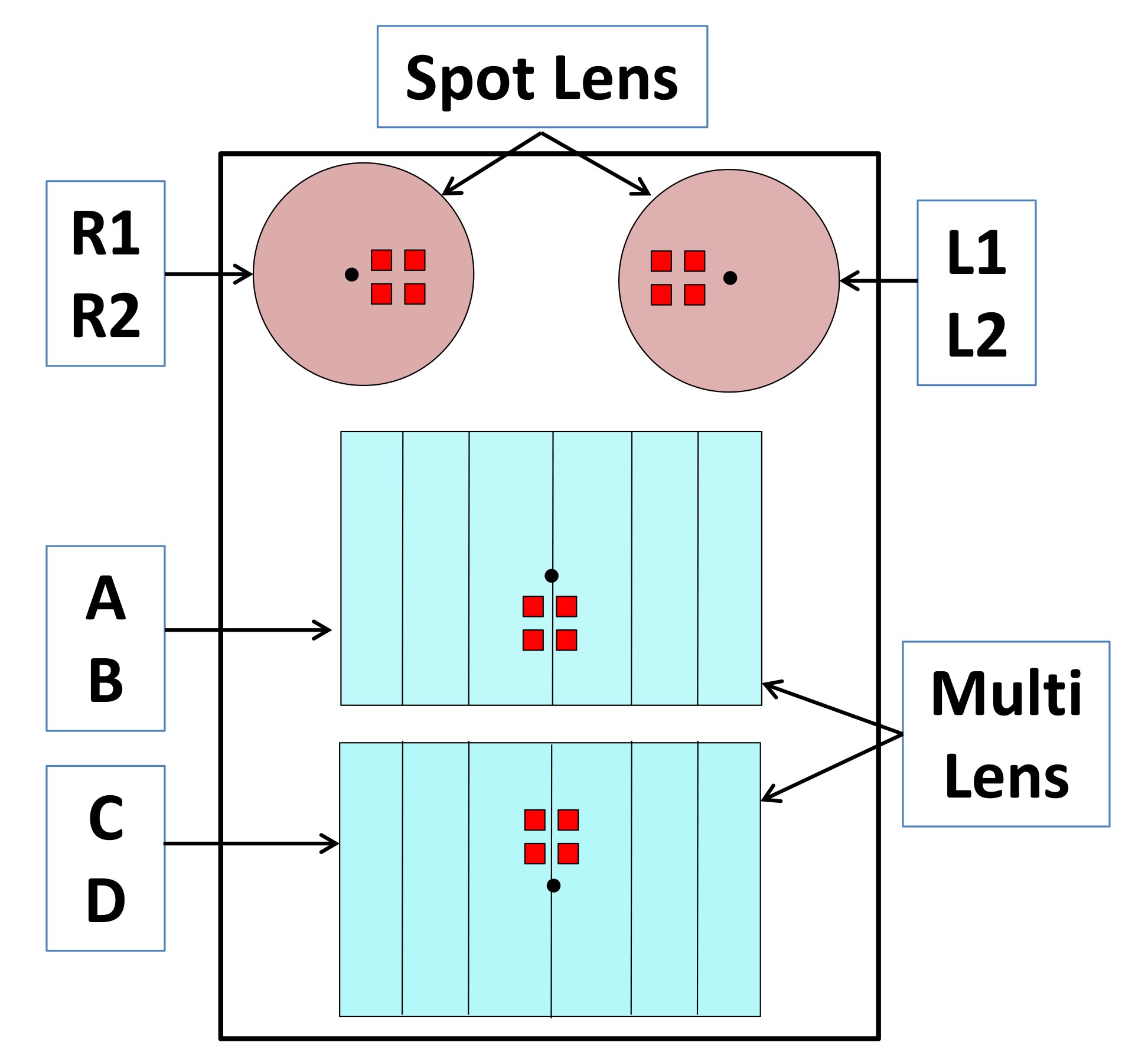}} 
\subfigure[VPA generated at a distance of 5m]{\label{fig:STP_VPA}\includegraphics[trim= 0.15in  0.75in 0.3in 0in, height=0.79in]{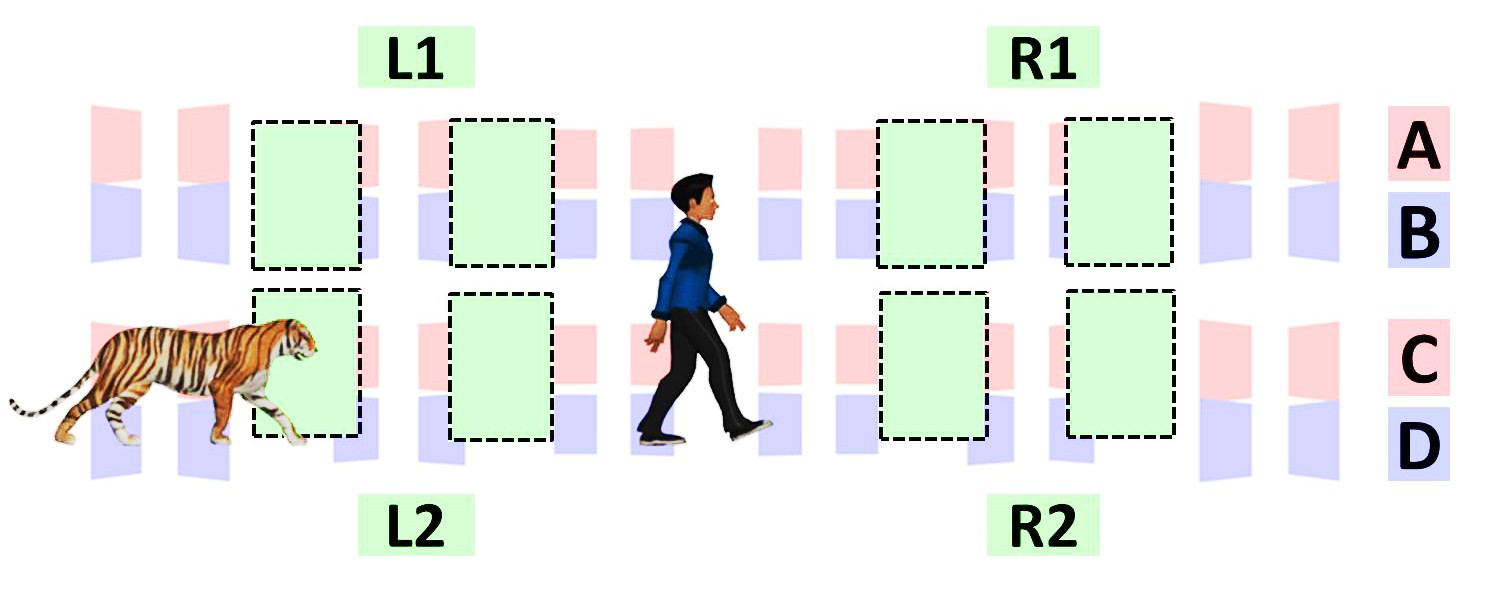}}
  \vspace{-0.15in}
\subfigure[Signals generated by human and animal moving across the VPA one followed by the other.]{\label{fig:8signals}\includegraphics[trim= 0.15in  0.75in 0.3in 0in, height=1.3in]{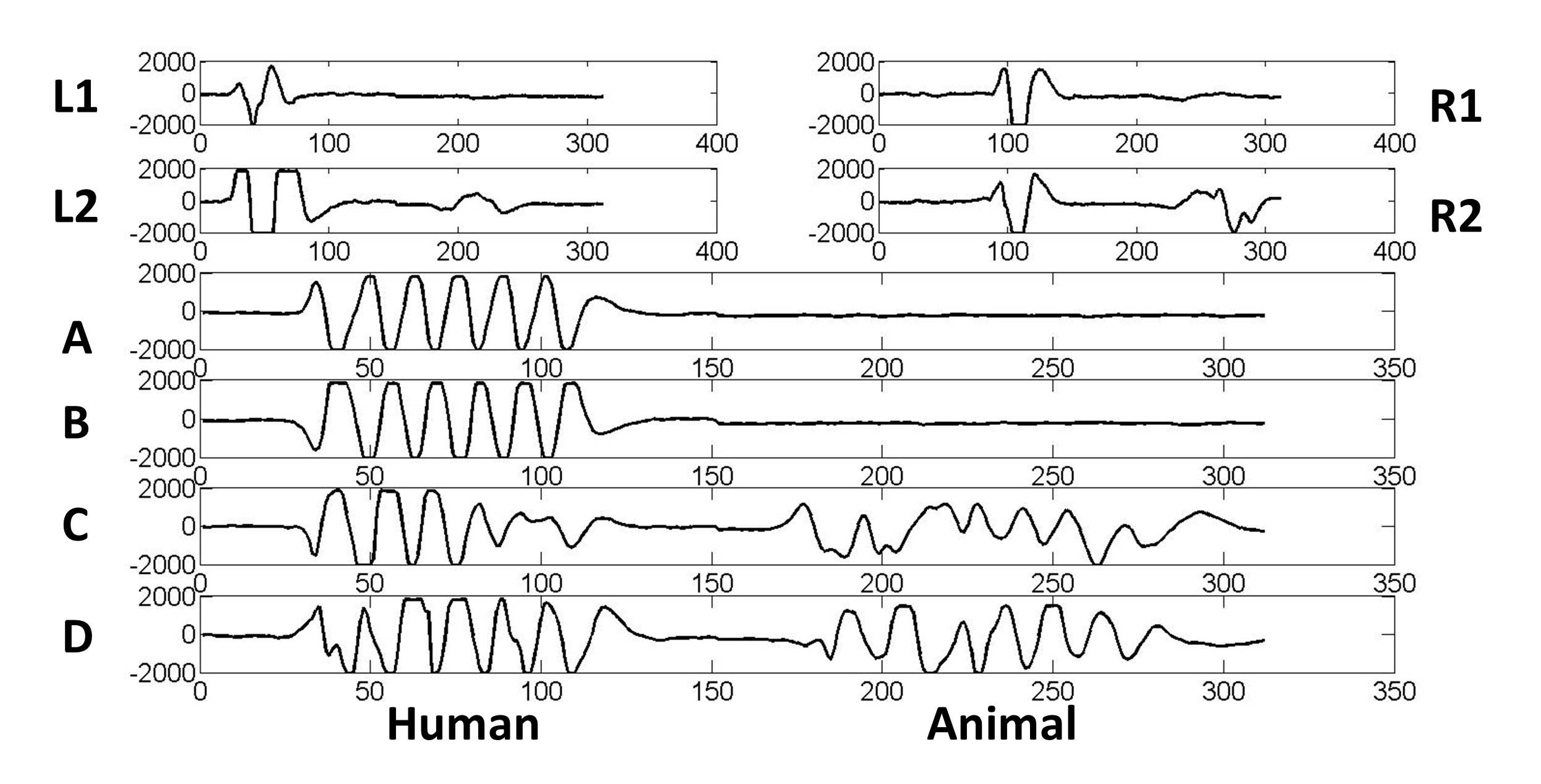}}
\caption{Illustrating the VPA design and typical associated signals for a human and animal motion.}
\end{figure}

The sensor platform consists of an array of $8$ sensors arranged in the form of a tower (and hence referred to as sensor tower platform), which are labelled as $A,B,C,D,L_1,L_2,R_1,R_2$ (see Fig.~\ref{fig:Final_Design}). Sensors $A,B,C$ and $D$ are arranged as a vertical array (see Fig.~\ref{fig:Final_Design}) to provide the spatial resolution needed to distinguish between human and animal based on their height. Sensors $A,B$ share a common multi-lens (called multi-lens $AB$) as do sensors $C,D$ (multi-lens $CD$). The VPA associated with sensors $A,B,C,D$ is shown in Fig.~\ref{fig:STP_VPA}. 

Sensors $L_1,L_2,R_1,R_2$ are placed so as to form two vertical arrays (see Fig.~\ref{fig:Final_Design}) to distinguish between translational and oscillatory motion. Sensors $L_1$, $L_2$ share a single spot lens $L$ as do sensors $R_1,R_2$ (spot lens $R$).    The VPA associated with sensors $L_i,R_i$, $i=1,2$, is shown in Fig.~\ref{fig:STP_VPA}.  If the intruder motion is translatory in nature with motion taking place from left to right of the STP, then the signal seen by sensor $R_i$ will in general be a delayed version of the signal seen by $L_i$.  This will not be the case with oscillatory motion in general and it is this feature that we exploit in distinguishing between intruder and clutter. 

\begin{figure}[h]
   \vspace*{-0.1in}
    \centering
	\includegraphics[trim= 0.15in  0.75in 0.3in 0in, width=2in]{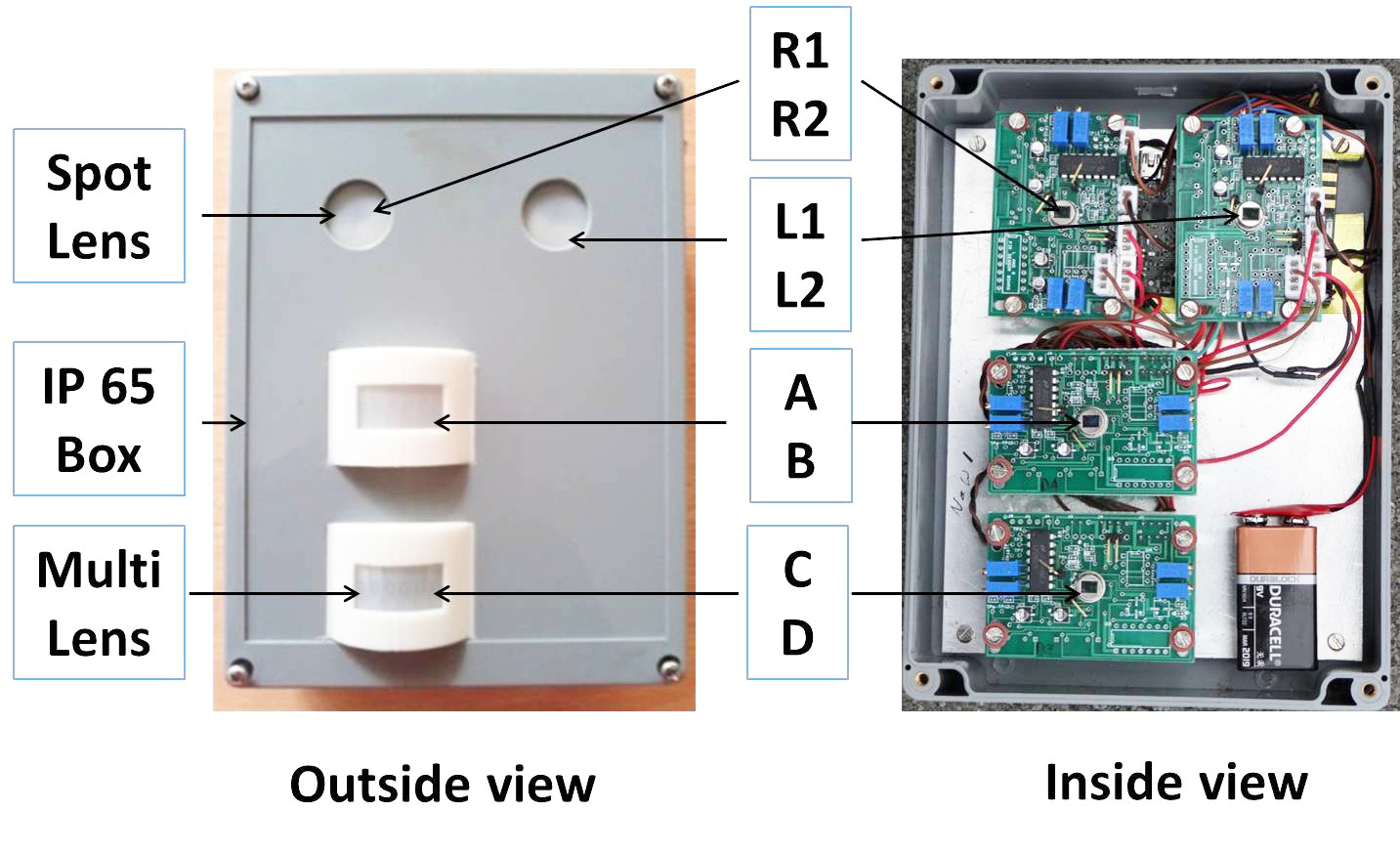}
	  \vspace{-0.1in}
    \caption{The STP developed and its associated electronics. \label{fig:STP}}
\end{figure}

The sensors and amplifier electronics are housed inside an IP65 box.  The lens mounts for the multi-lenses $AB$ and $CD$ were made using a 3D printer.   A snapshot of the developed STP appears in Fig.~\ref{fig:STP}. For additional detail relating to components used\footnote{The REP05B PIR sensor package has its pixels arranged in a $2\times 2$ fashion. The sensor provides two outputs, one for each row of differentially wired pixels.}, please see Table~\ref{tab:component}.
  \vspace*{-0.2in}
\begin{center}
	\begin{table}[h]
		\begin{center}
			\caption{Sensors and lenses employed in each STP. } 		\label{tab:component}
			\begin{tabular}{||c|c|c|c||} \hline \hline 
			Component	& Quantity  & Company & Part No. \\ \hline 
				PIR Sensor & 4 & Nippon Ceramic  & REP05B  \\ \hline 
				Multi Lens&  2 & Kube Electronics &TR426	  \\ \hline 
				Spot Lens&  2 & Kube Electronics &TR1004	  \\ \hline 
			\hline 
			\end{tabular}
		\end{center}
	\end{table}
\end{center}
\subsection{Use of Vertical Offset to Avoid Overlap Between Rows of the VPA}
\begin{figure}[h]
  \centering
  \vspace*{-0.15in}
  \subfigure[]{\label{fig:overlap}\includegraphics[trim= 0.15in  0.9in 0.3in 0in, height=1in]{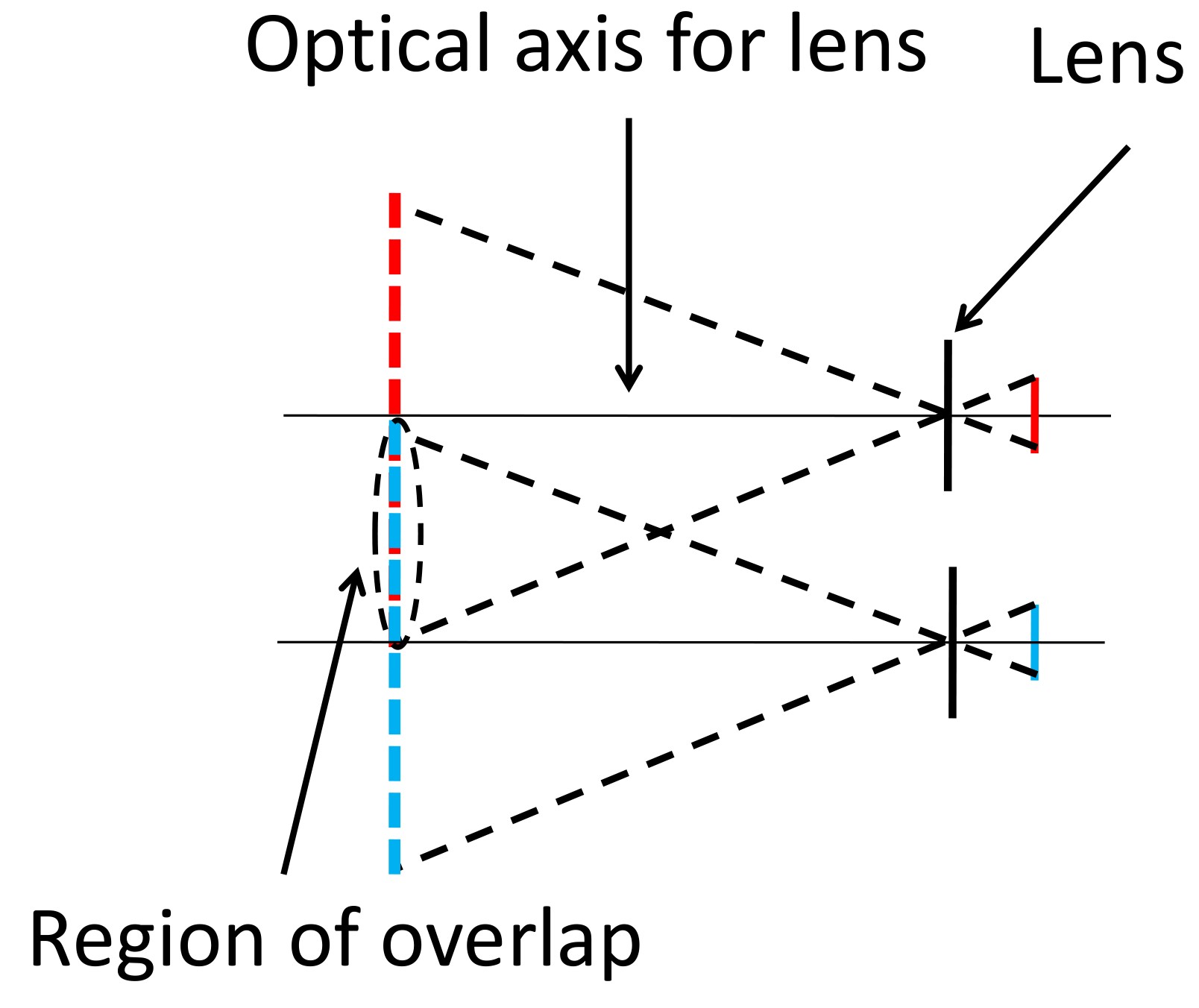}}
 \hspace{1cm}
      \subfigure[]{\label{fig:vertical_offset}\includegraphics[trim= 0.15in  0.5in 0.3in 0in, height=1in]{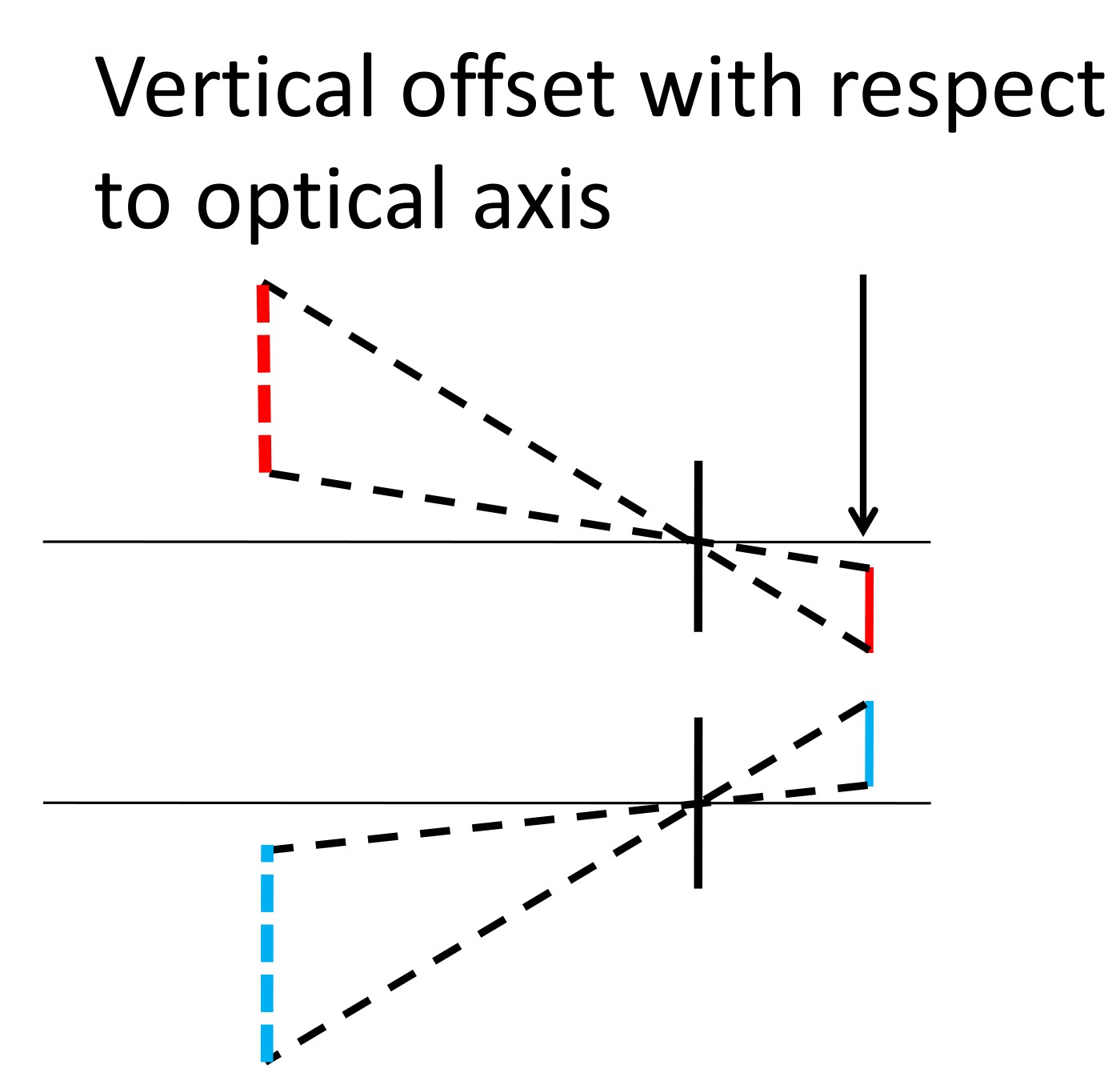}}
      \vspace*{-0.1in} 
   \caption{ In (a), it is shown that close spacing of two lenses (equivalently, multi-lenses) with sensors centered on the optical axes of the two lenses causes beams to overlap at a distance.   This is prevented by the simple arrangement of sensors in (b), where the sensors are vertically offset with respect to the optical axes of the corresponding lenses.     \label{fig:offset}}
 \end{figure}

The need for making the STP as compact as possible, forces the close placement of the multi-lenses $AB$ and $CD$.  However, close placement of the two multi-lenses will cause rows $B$ and $C$ of the VPA to overlap at larger distances from the STP, leading to a loss in spatial resolution.  To overcome this, we resorted to the simple but effective trick of vertically offsetting sensors $A,B$ and $C,D$, i.e., placing sensors $A,B$ slightly below the optical axis of multi-lens $AB$ and sensors $C,D$ slightly above the optical axis of multi-lens $CD$.  This is illustrated in Fig.~\ref{fig:offset}.   
\subsection{Electronics for Sensor Platform}
\begin{figure}[h]
   \vspace*{-0in}
    \centering
	\includegraphics[trim= 0in  0in 0in 0in, width=2.75in]{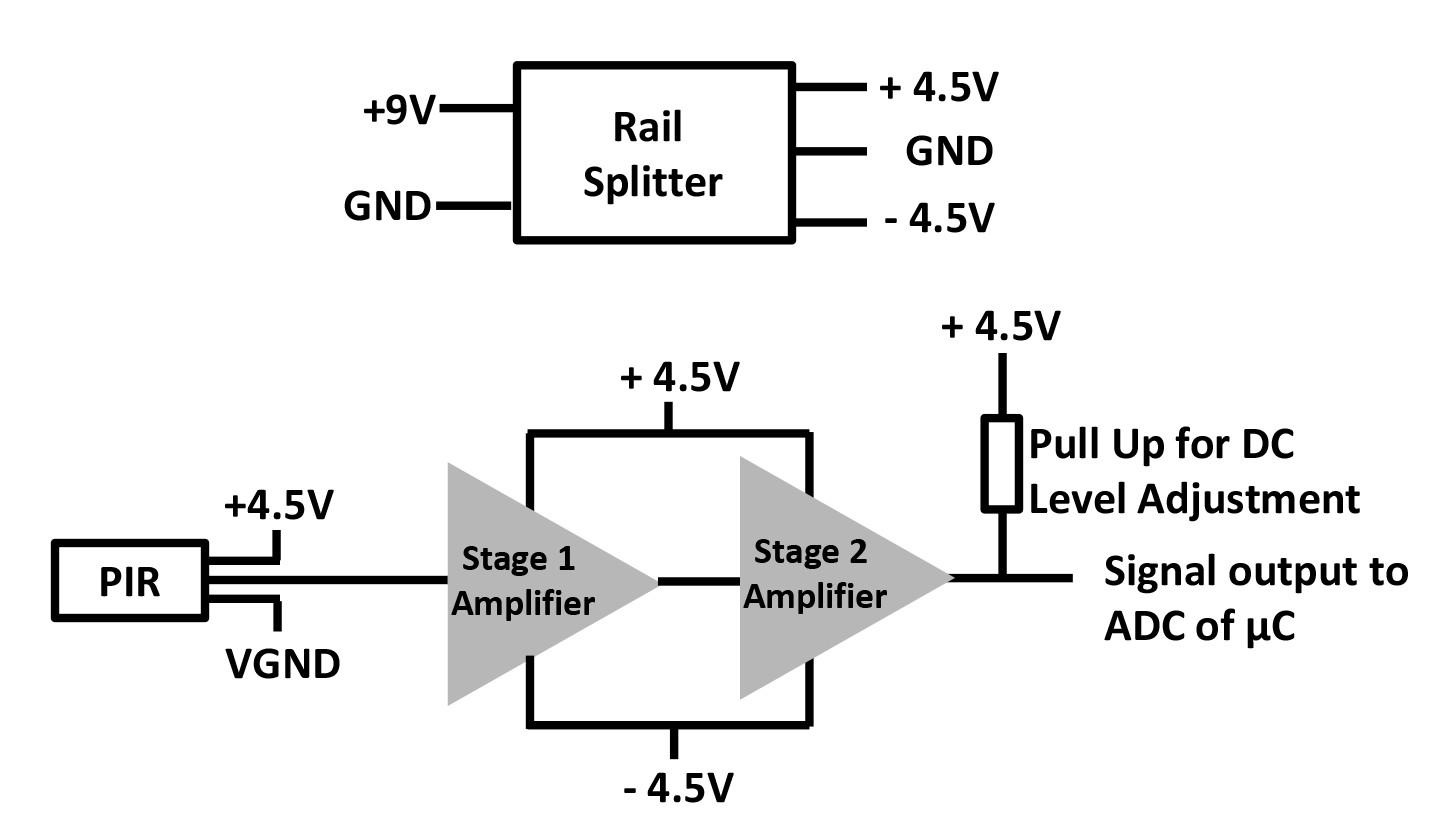}
	  \vspace{-0in}
    \caption{Two stage amplifier electronics for PIR Sensor. \label{fig:AmpDesign}}
\end{figure}

The output of the PIR sensor is weak (typically on the order of $\mu$V) and needs amplification. A two-stage amplifier circuit is used  to boost the signal amplitude (shown in Fig. \ref{fig:AmpDesign}). Potentiometers are used as feedback resistors at each amplifier stage to provide a variable gain. The gain is adjusted in order to maintain a signal with large amplitude but free from saturation.  The reason for this is that the signal processing techniques we employ for discriminating between various classes, is dependent on variations in signal amplitude. In order to improve the dynamic range of the amplifier, we provide a dual supply to the amplifier. This necessitates the inclusion of a rail-splitter. The rail-splitter takes the conventional single rail from a $9$ V DC supply and splits it to provide a $\pm$ $4.5$ V supply to the amplifiers.  The output of the final amplification stage is DC level shifted by using a pull-up resistor in order to make the polarity of the signal positive (as the analog to digital converter on the mote requires signal inputs with a positive polarity only).
 
 
The PIR sensor signal is a function of the relative temperature difference between the moving object and its background (see \eqref{eq:radiation}). The larger the temperature difference the larger will be the amplitude of the PIR sensor output and vice versa. Thus, a fixed gain cannot be used in a practical setting where the background temperature varies throughout the day. To address this limitation one could measure the ambient temperature and adjust the gain of the potentiometers appropriately. 

\begin{figure}[h]
   \vspace*{-0in}
    \centering
	\includegraphics[trim= 0in  0in 0in 0in, width=2.5in]{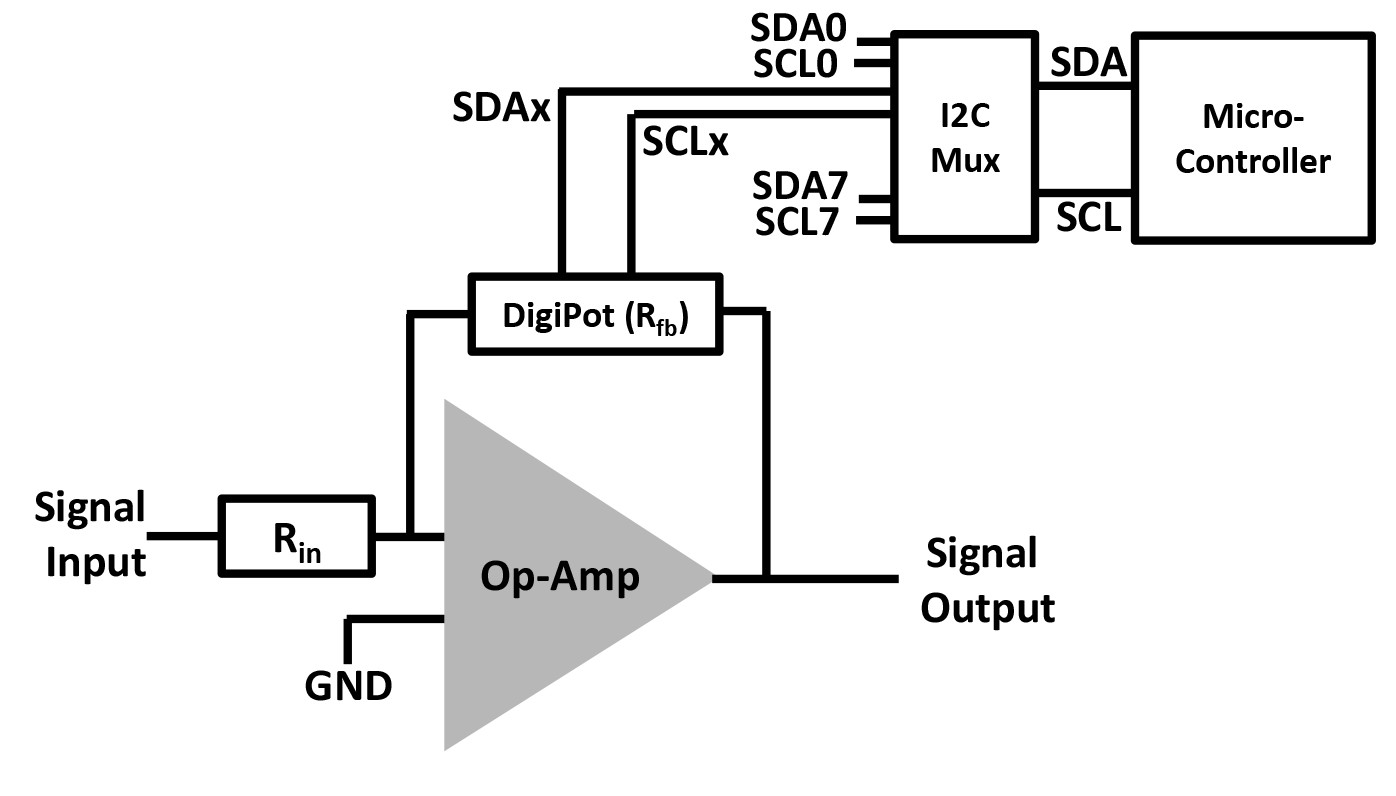}
	  \vspace{-0in}
    \caption{Digital potentiometer for  automatic gain control. \label{fig:digipot_electronics}}
\end{figure}

In order to avoid the cumbersome and time-consuming task of manually adjusting the potentiometers, we replace them with digital potentiometers whose resistance can be adjusted using a micro-controller (Fig. \ref{fig:digipot_electronics}). An inter-integrated circuit (I2C) driver is used in conjunction with an I2C multiplexer to program each amplifier stage. We plan to utilize this circuit in conjunction with an ambient temperature sensor for automatic gain adjustment in future experiments. 

\subsection{Operating Range of the STP}

\begin{figure}[h]
    \centering
    \vspace*{-0.15in}
	\includegraphics[trim= 0.15in  0.25in 0.3in 0in, width=2in]{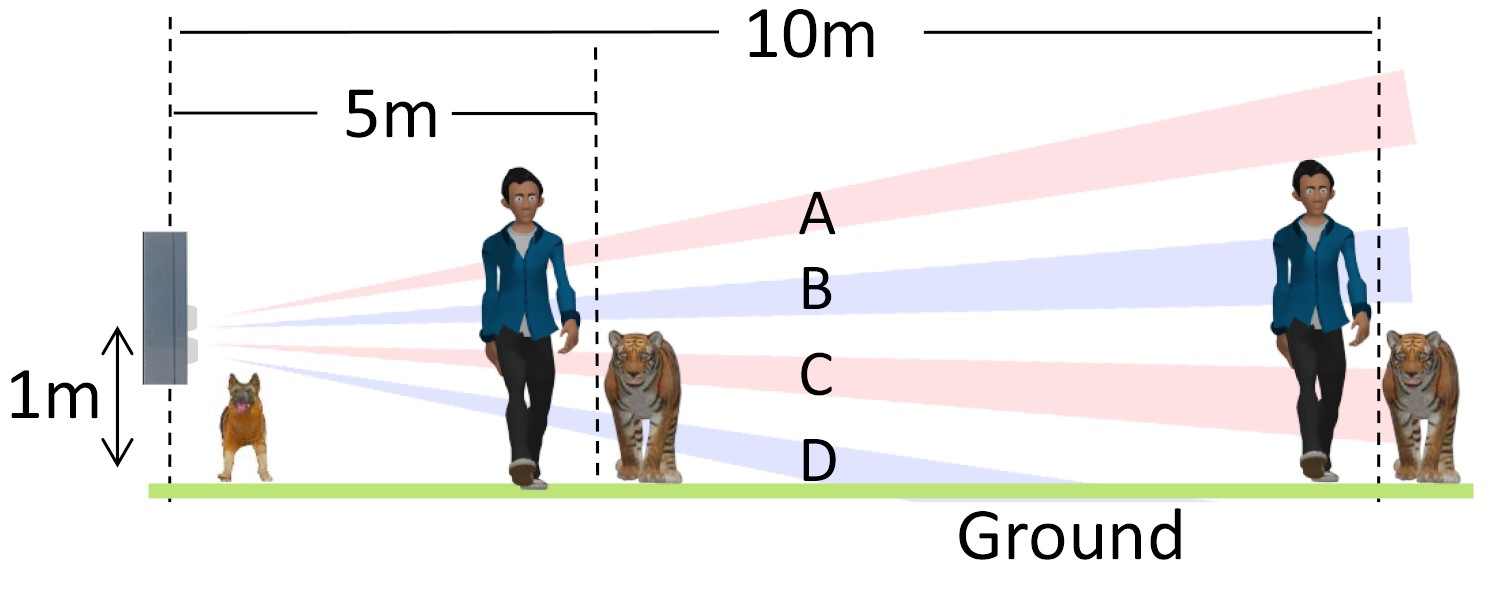}
	\vspace*{-0.05in}
    \caption{The effect of diverging nature of the VPA. \label{fig:ClassificationRange}}
\end{figure}

The STP is required to have the ability to distinguish between human and animal over a range of distances.  A side view of the FoV of the STP is shown in Fig.~\ref{fig:ClassificationRange}.  The inherent sensitivity of the PIR sensors as well as the divergent nature of the beams, making up the FoV, limit the operating range of the STP.    The operating range of the STP developed here is 5-10m.   At distances smaller than $5$m, a small animal such as a dog will pass below all $4$ beams $A,B,C,D$ and hence will travel undetected as can be seen from Fig.~\ref{fig:ClassificationRange}.   At ranges beyond $10$m, the sensitivity of the sensors employed is not sufficient to guarantee detection with the required accuracy.  
\vspace*{-0.15in}
\section{Data Collection}  \label{sec:data_collection} 
\begin{figure}[h]
    \centering
    \vspace*{-0.15in}
         \subfigure[]{\label{fig:inclined_walk}\includegraphics[trim= 0in  0.25in 0in 0in, width=1.4in]{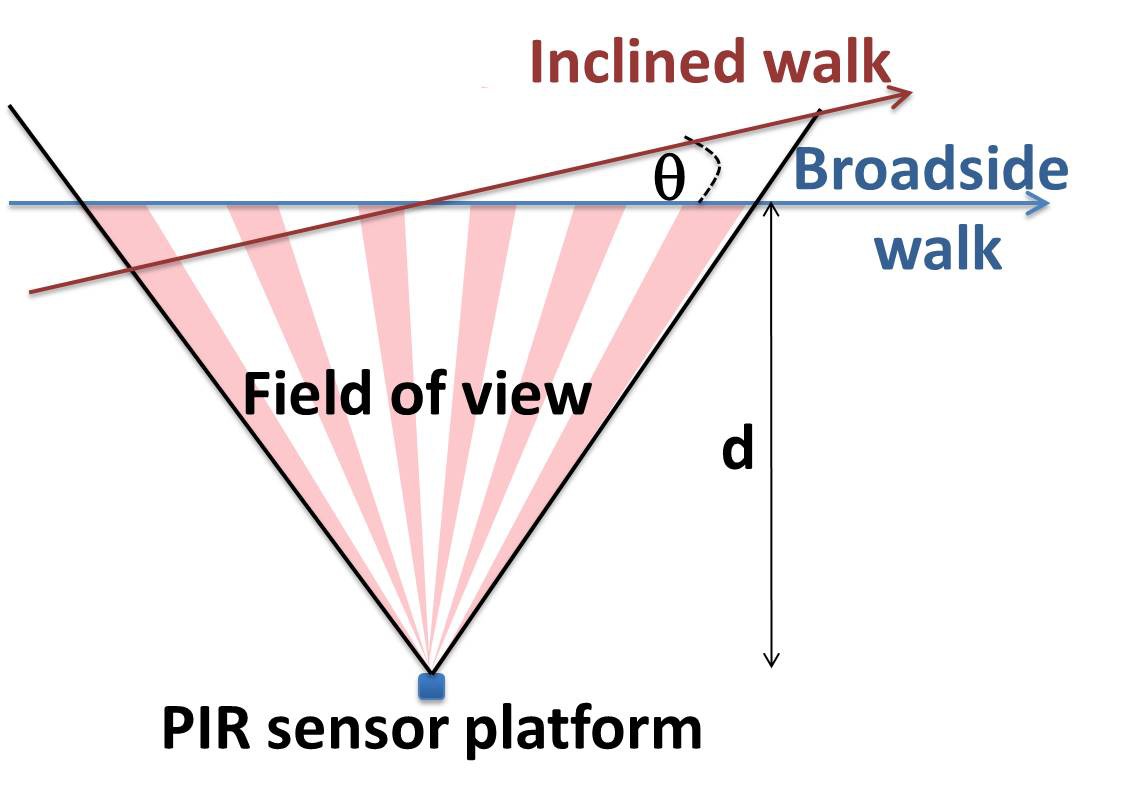}}
           \hspace{0cm}
                \subfigure[]{\label{fig:LeopardDataCollection1}\includegraphics[trim= 0in  0.25in 0in 0in, height=0.8in]{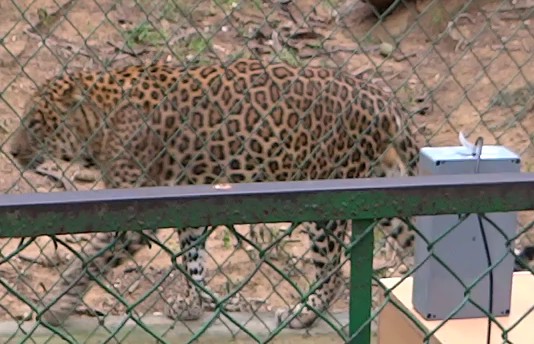}}
                \vspace*{-0.1in} 
       \caption{(a) The angle of inclination $\theta$ of an intruder path relative to broadside walk is illustrated here.    (b) One of the settings in which data were collected in the Bannerghatta Biological Park.   The STP collecting data is visible in the foreground to the right.   \label{fig:dataCollecction}}
\end{figure}

A key step in supervised machine learning is data collection.   Supervised classification algorithms require labelled data from which the machine learns the classification model.  The $3$ classes of interest here are  humans, animals and clutter.   In general, data collection in the case of animals, particularly wild animals, is both challenging as well as time consuming.   We collected dog-intrusion data at a dog-trainer facility in Bengaluru.  Data corresponding to tiger, leopard and wolf intrusions were collected at the Bannerghatta Biological Park (BBP).  Data corresponding to human intruder motion as well as clutter were gathered on the forested campus of the Indian Institute of Science.  The collected data corresponded to straight-line motion at a variety of typically observed speeds and angles of inclination $\theta$ (see Fig.~\ref{fig:inclined_walk}) at various distances within the desired operating range of $5$-$10$m. 
\section{ASPIRE: Animation-based Simulation tool for Passive Infra-Red sEnsor} \label{sec:animation} 

 \begin{figure}[h]
    \centering
	\includegraphics[trim= 0.15in  0.25in 0.3in 0in, width=3.25in]{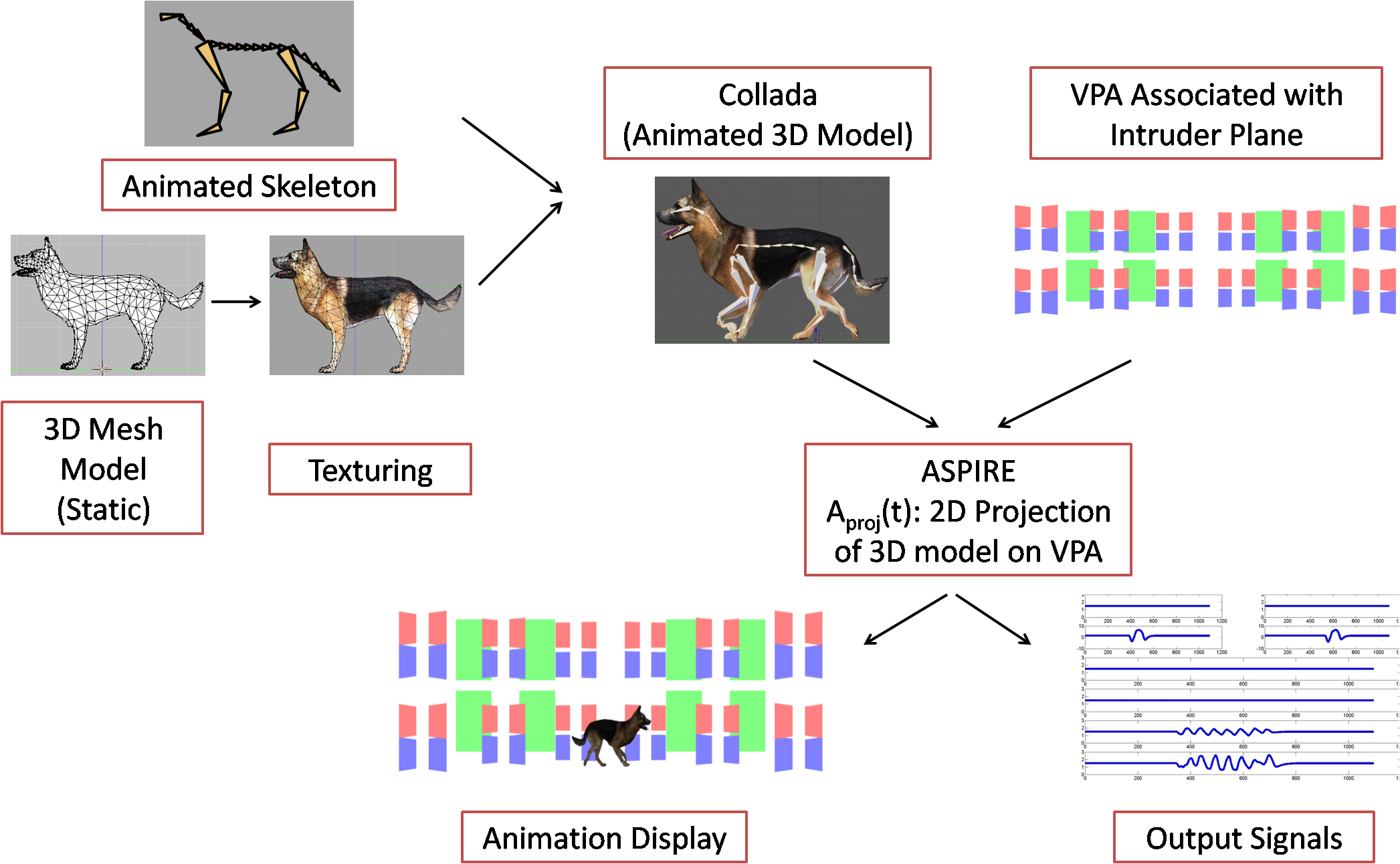}
    \caption{ASPIRE block diagram to obtain the simulated PIR signal. \label{fig:ASPIRE_BlockDiagram}}
\end{figure}

Given the difficulty noted in Section~\ref{sec:data_collection}  in gathering data associated to animal motion, an Animation-based Simulation tool for Passive Infra-Red sEnsor (ASPIRE) was developed to simulate data corresponding not only to animal motion, but also human motion and clutter as well. Animated 3D models of the intruder were developed using the popular animation software Blender.  The 3D model consists of a collection of triangles that are arranged in the form of a mesh that approximates the outer surface of the intruder. 

The key intermediate step in determining the signal resulting from motion of the 3D model across the FoV of the STP is to find the area of the 3D mesh when projected onto the VPA of the STP.  By VPA here, we mean the VPA that is associated with the vertical plane along which the intruder moves.   The area of projection of the 3D mesh can in turn be determined by projecting all triangles in the mesh model onto the VPA and computing the area of the polygon obtained by taking the union of all these projected triangles. 

\begin{figure}[h]
  \centering
  \vspace*{-0.15in}
  \subfigure[]{\label{fig:calculating_waveforms_b}\includegraphics[trim= 0.0in  0.5in 0.0in 0in, height =0.6in]{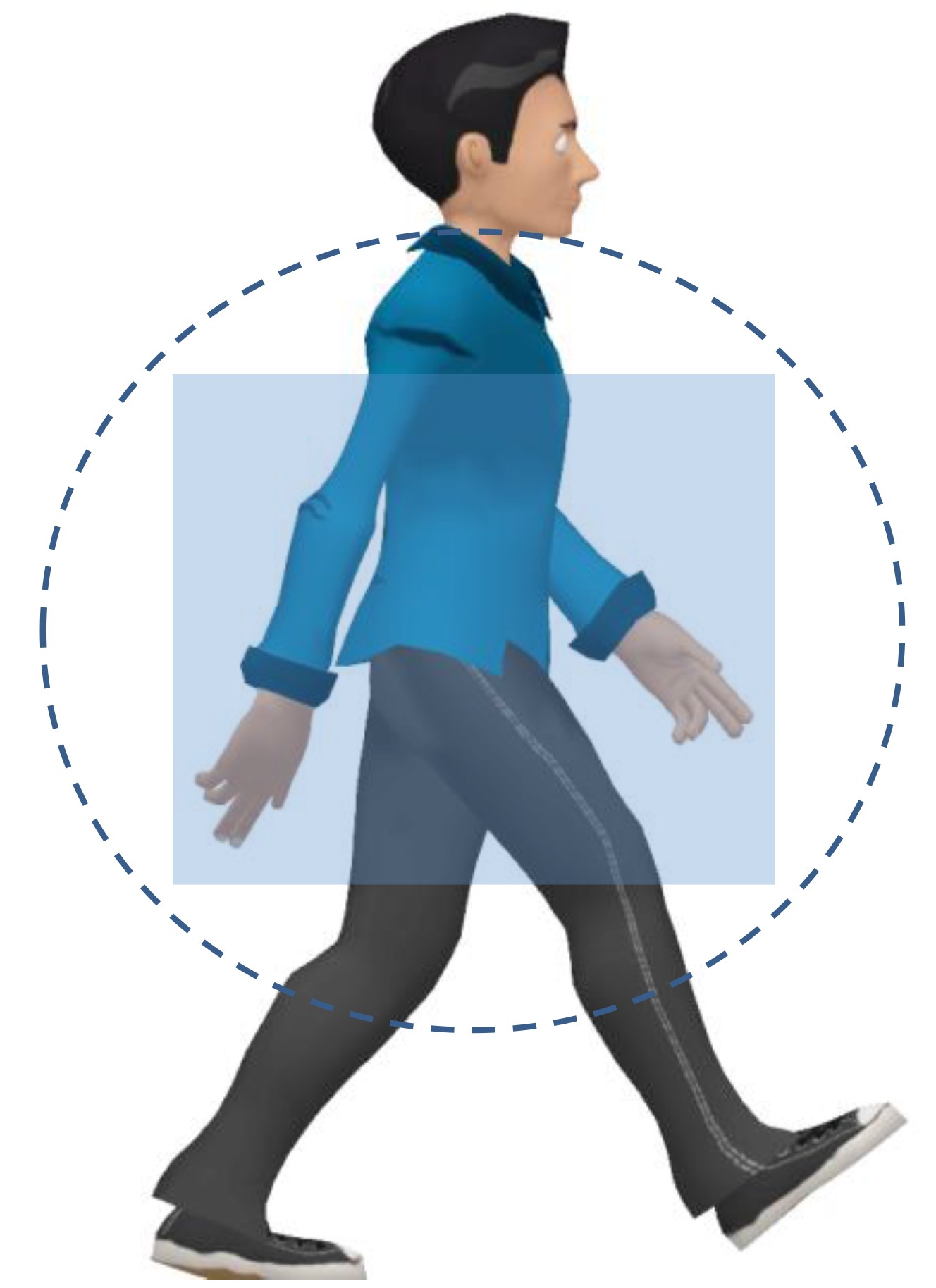}}
      \hspace{1.5cm}
  \subfigure[]{\label{fig:calculating_waveforms_c}\includegraphics[trim= 0.0in  0.25in 0.0in 0in, height =0.5in]{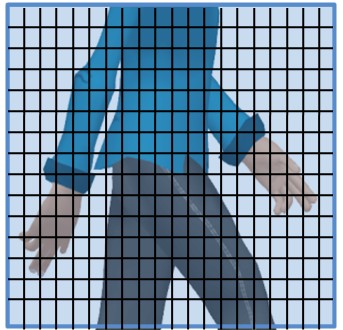}}
  \vspace*{-0.1in} 
  \caption{Calculating the area of the intruder when projected onto the VPA.}
\end{figure}

To simplify computation, in place of actual area computation, what is done in practice is to divide the VPA into a very large number of tiny squares. The area is then to a good approximation equal to be the sum of the areas of all squares that have non-empty intersection with the projected 3D mesh.  This is illustrated in Figs.~\ref{fig:calculating_waveforms_b} and \ref{fig:calculating_waveforms_c}.  This approach is along the lines of the approach employed in computer graphics for rendering a scene~\cite{ZBufferAlgorithm}.

Animation techniques have been used previously to simulate waveforms in conjunction with doppler sensing for the purposes of detection and classification of different types of human motion\cite{ram2008simulation}.  In that paper, the authors use Bio-Vision Hierarchy (BVH) files to simulate the motion of a human being as opposed to the more sophisticated Collada file format that we developed using Blender software.  

\section{Chirplet-Based Model for Intruder Signal} \label{sec:chirp} 
In the classification algorithm that is discussed in Section~\ref{sec:results}, we exploit the simple-but-useful observation, made to our knowledge for the first time here, that intruder signals are well modeled using a chirp.  More specifically, the signals generated by the intruder moving across $A,B,C,D$ arrays of the VPA exhibit chirp.  

\begin{figure}[h]
    \centering
 \vspace*{-0.05in}
	\includegraphics[ trim= 0in  0.5in 0in 0in, width=2in]{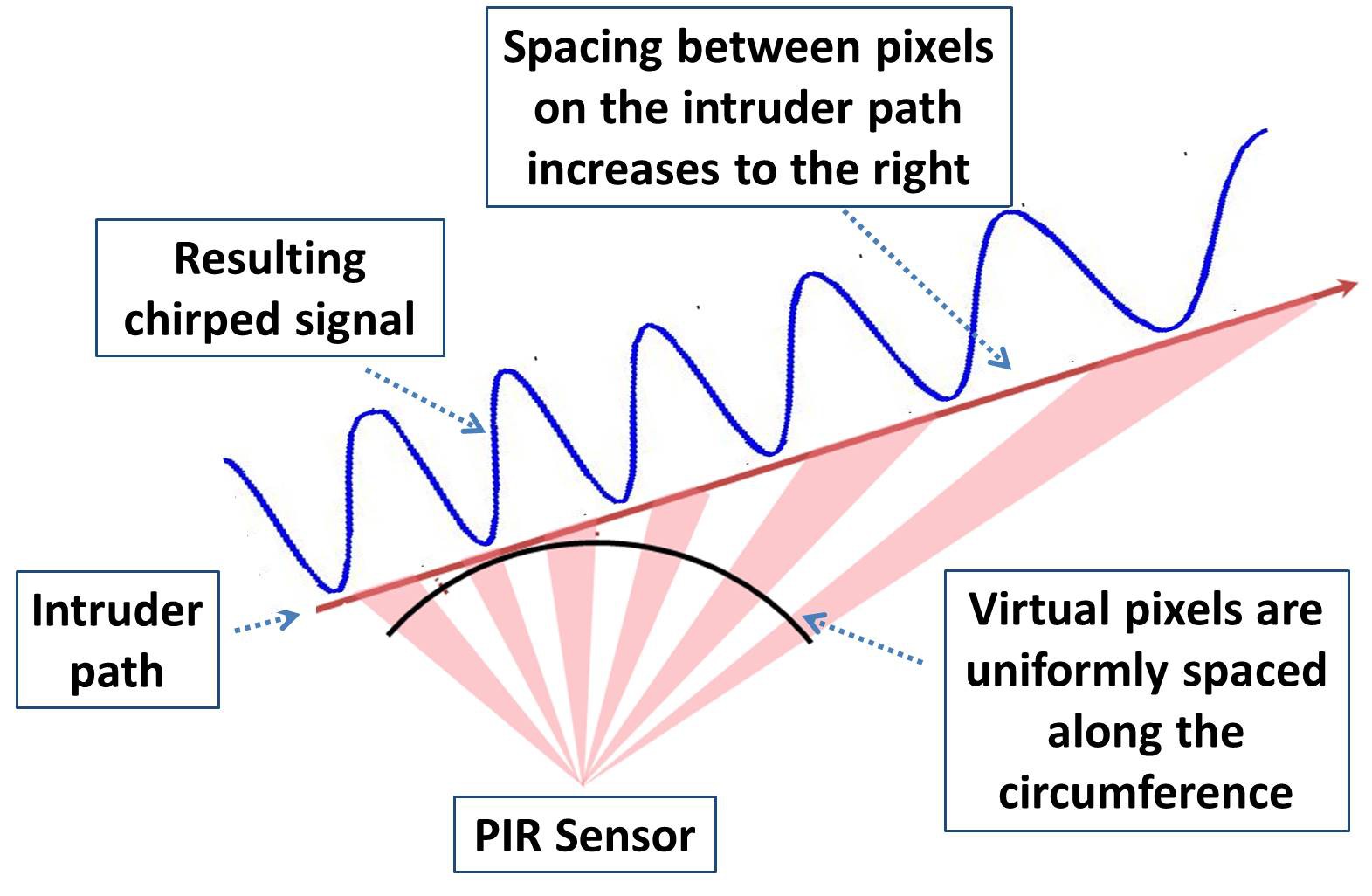}
    \caption{Explaining the chirped signal that arises from intruder motion along a straight line at uniform velocity.  \label{fig:chirp_explanation2}}
\end{figure}

\begin{figure}[h]
  \centering
  \vspace*{-0.2in}
  \subfigure[]{\label{fig:UpChirp_Example}\includegraphics[trim= 0in  2.25in 0in 0in, height=0.35in, width =1.5in]{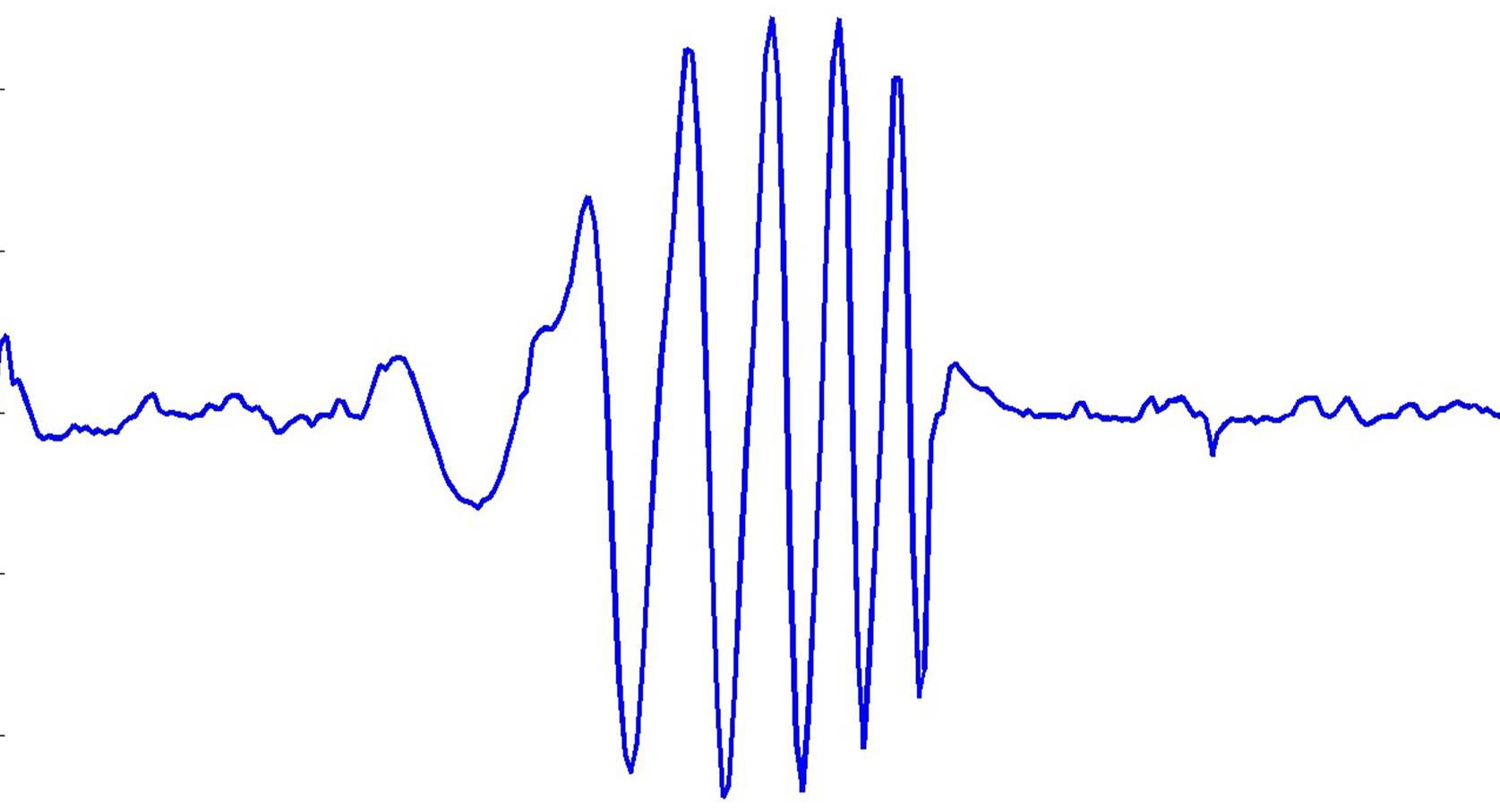}}
    \hspace{0.5cm}
  \subfigure[]{\label{fig:Clutter_ChirpExample}\includegraphics[trim= 0in  2.25in 0in 0in, height=0.5in, width =1.5in]{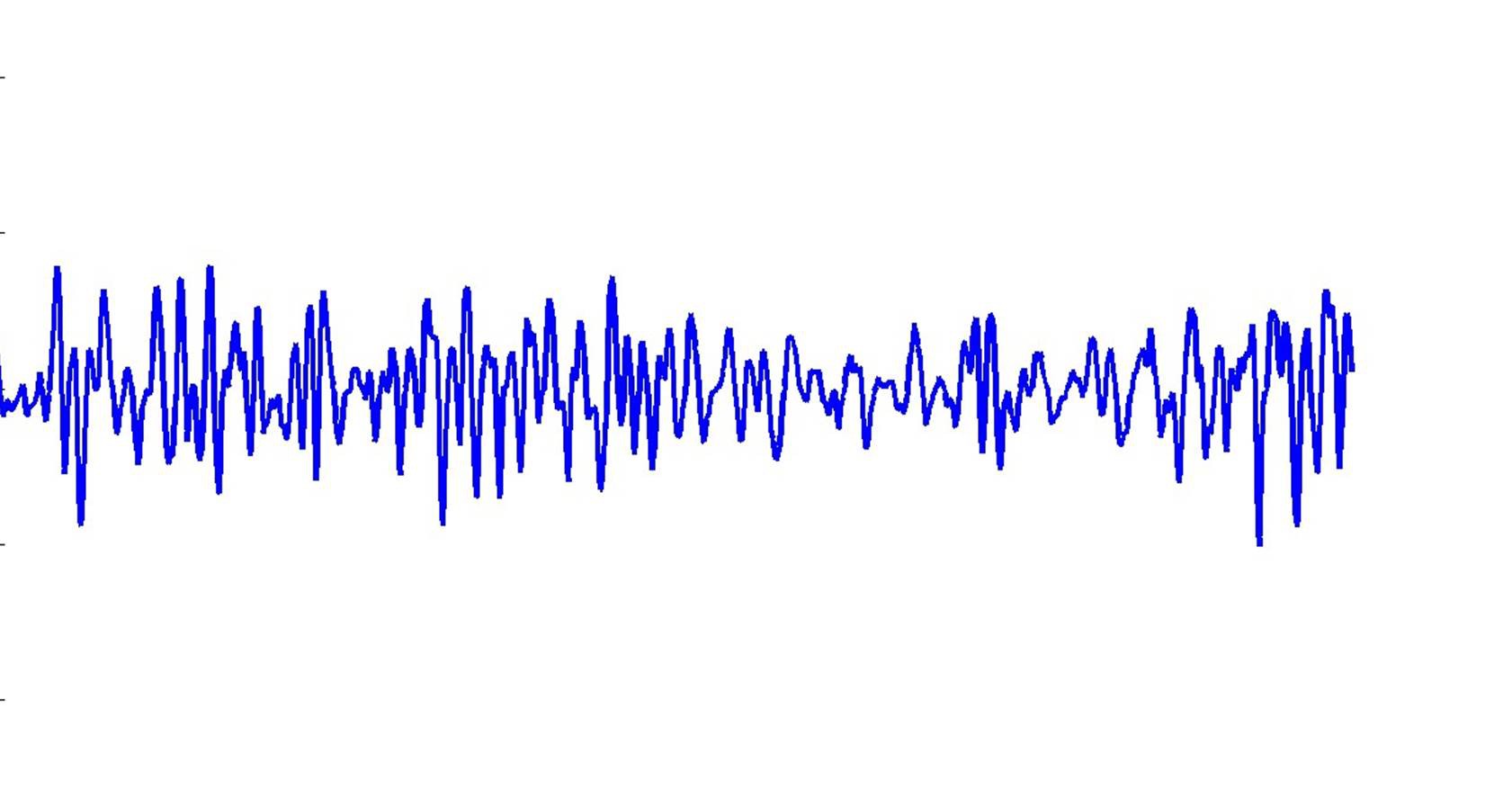}}
  \vspace*{-0.1in}
  \caption{Illustrating intruder and clutter signals.}
\end{figure}

The chirp phenomenon is explained in Fig.~\ref{fig:chirp_explanation2}.  This figure presents a top view of the virtual beams generated by employing a multi-lens in conjunction with a PIR sensor.
Consider an intruder moving along a circular path at uniform speed.  In this case, the size of the virtual pixels encountered and the gaps between two consecutive pixels of the VPA will be equal throughout the duration of intruder motion.  As a result, the response of the PIR sensor will resemble a periodic sinusoidal signal.  In the case of intruder motion from left to right along the straight-line path shown in Fig.~\ref{fig:chirp_explanation2}, the size of the virtual pixels and gaps between two successive pixels of the VPA initially decreases until the intruder reaches the point of closest approach and thereafter increases. Thus, an intruder moving at a uniform speed along such a straight-line path will generate a signal comprising of an up-chirp followed by a down-chirp as illustrated at the top of Fig.~\ref{fig:chirp_explanation2}.   The extent and nature of the chirp will depend upon the angle $\theta$ of inclination of the intruder path. 
Clearly, in the case of oscillatory motion corresponding to clutter, no such chirp will be present.   Figs.~\ref{fig:UpChirp_Example} and \ref{fig:Clutter_ChirpExample} show signals of intruder and clutter events, respectively.
\vspace*{-0.13in}
\subsection{Intruder Detection via Chirplet Decomposition}
Our decomposition of the signal waveform as a sum of chirplets makes use of complex-exponential representation of elemental chirped signals in the form: 
\bea
x(n; m, \omega, c, d) \ = \  ({2\pi}d^2)^{-\frac{1}{4}}\exp {\bigg \{-\bigg (\frac{n-m}{2d}\bigg )^2 \bigg \} }  \nonumber \\ 
\times \exp {\bigg \{j\omega(n-m)+j\frac{c}{2}(n-m)^2}\bigg \}, \label{eqn:Chirplet}
\eea
where, $m$, $\omega$, $c$ and $d$ are the parameters of the chirp signal representing the location in time, location in frequency, chirp rate and duration, respectively. 

As a first step, we pass on from the real intruder signal $s(n)$ to its complex, analytic representation $s_a(n) \ = \ \left(s(n) + j \hat{s}(n)\right)$, where $\hat{s}(n)$ is the Hilbert transform of $s(n)$.  We next decompose $s_a(n)$ as the weighted sum of $q$ chirplets 
\beq
s_a(n)= \sum_{i=1}^q  a_ie^{j\phi_i}x_i(n; m_i, \omega_i, c_i, d_i),
\eeq
where, each $x_i(n)$ is given by an expression as in \eqref{eqn:Chirplet}. 

\begin{figure}[h]
   \begin{center}
   \includegraphics[trim= 0in  0.25in 0in 0in, width=3in]{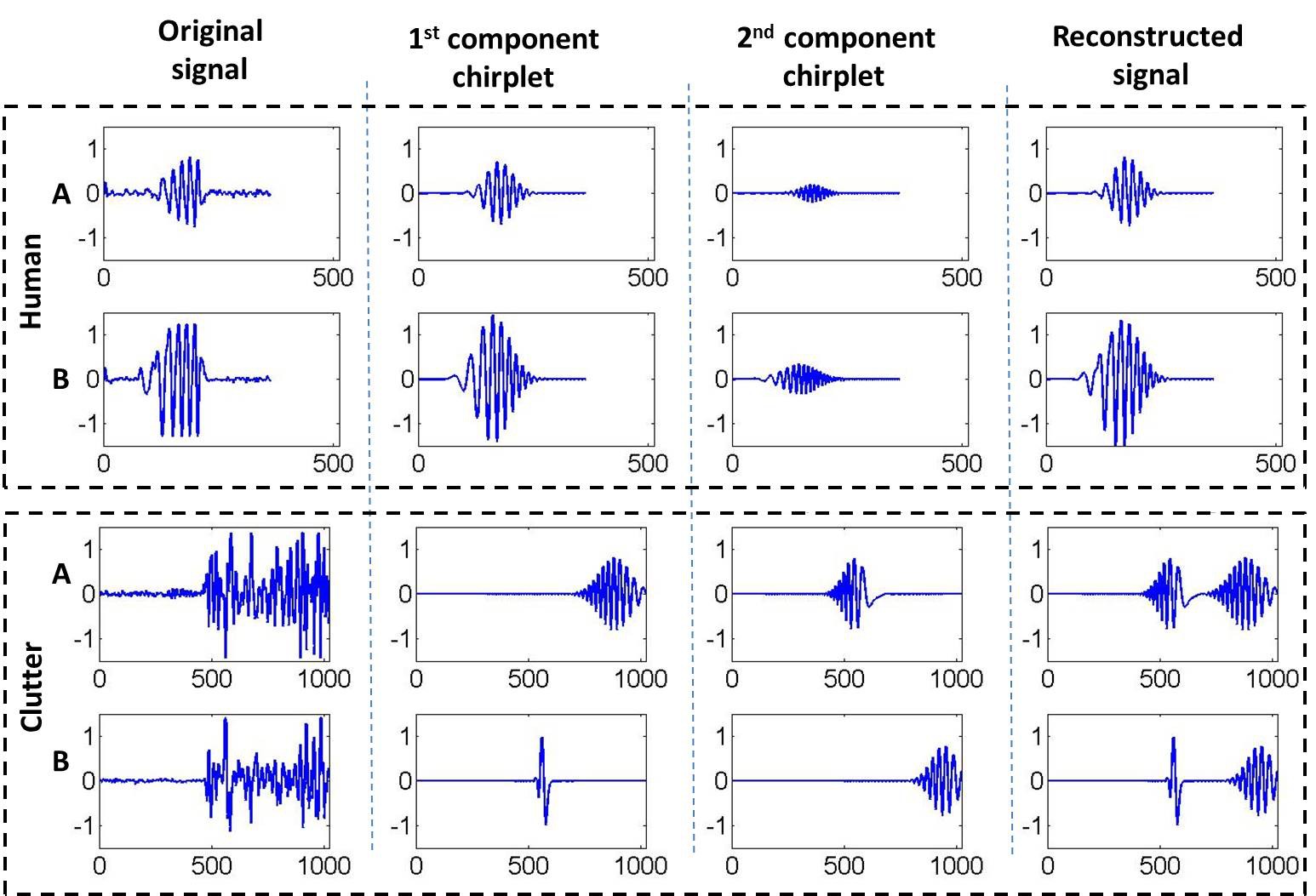}
   \caption{ Chirplet decomposition of the signals associated to sensors $A$ and $B$ in the case of both intruder motion as well as  clutter are shown here.  For the sake of clarity, only two components chirplets are shown.}
   \label{fig:chirp_explanation}
   \end{center}
    \end{figure}

Next, maximum likelihood estimates of the parameters $a$, $m$, $\omega$, $c$ and $d$ of each chirplet are found using the method in \cite{o2000sparse}. The intruder signals turn out to be well approximated by the sum of three chirplets. The chirplet decomposition and subsequent reconstruction of an intruder and clutter signal associated to sensors $A$ and $B$ are shown in Fig.~\ref{fig:chirp_explanation}.   

Note that the reconstructed signal is close to that  of the original signal in the case of an intruder signal and  this is not true in the case of signal arising from clutter.    What is fed to the SVM, however, are the parameters corresponding to this $3$-chirplet decomposition.  As will be seen in Section~\ref{sec:results}, this approach resulted in high classification accuracy for the data sets collected.  Some justification for this approach can be seen from the fact that in the chirplet decomposition associated to clutter, the times of arrival $m_i$, of the different chirplet components, are well separated in time unlike in the case of an intruder signal, where the signals arrive synchronously.  Also in terms of chirplet signal duration $d_i$, the different chirplets can have time durations that are very different, again unlike in the case of an intruder signal, where the time durations of the different chirplet components are very close to one another. 

\section{Other Features Used in Classification} \label{sec:other_features} 
Key components of the feature vector used for classification were the parameters obtained from chirplet decomposition.   Other important features employed relate to cross-correlation between signals obtained from sensors $L_1,L_2,R_1,R_2$ and the energy of signals from all $8$ sensors as explained below. 

\subsection{Features Related to Signal Cross-Correlation}
When an intruder moves say from left to right, the signals generated by sensors $L_1$, $R_1$ are slightly delayed and time-scaled versions of each other.  A similar statement is true in the case of signals generated by sensors $L_2$, $R_2$.   As a result, these signals will be highly correlated.  Clutter generated by vegetation whose motion is oscillatory as opposed to translational will be unlikely to exhibit high correlation.  

The precise quantity relating to signal correlation employed as part of the feature vector is the maximum cross-correlation parameter $\rho_{\max}$ corresponding to signals $s_{L_i}$, $s_{R_i}$ generated by sensors $L_i$, $R_i$, $i=1,2$  and given by:\\
\beq
\rho_{\max} = \max_k \sum_n \sum_i \frac{s_{L_i}(n+k)s_{R_i}(n)}{\sqrt{(E_{L_1}+E_{L_2})(E_{R_1}+E_{R_2})}},
\eeq
where, 
$E_{L_i} = \sum_{n}s_{L_i}^2(n)$ and $E_{R_i} = \sum_{n}s_{R_i}^2(n)$ for $i=1, 2$.

\begin{figure}[h]
  \centering
\subfigure[Real-world data]{\label{fig:Correlation_fig_Real_Data}\includegraphics[trim= 0in  1in 0in 0in, width=1.5in] {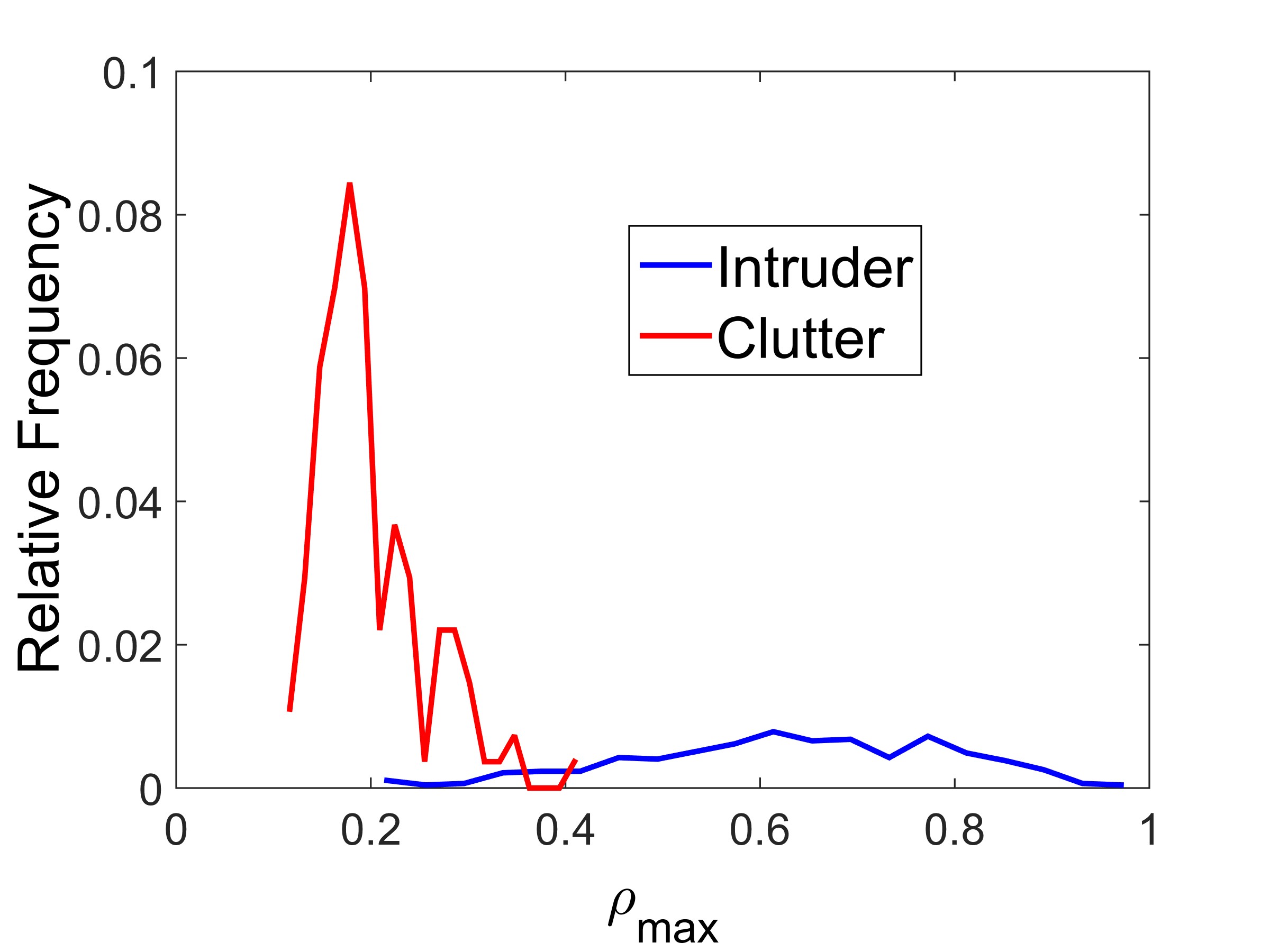}}
  \hspace{0.1cm}
\subfigure[Simulated data]{\label{fig:Correlation_fig_Anim_Data}\includegraphics[trim= 0in  1in 0in 0in, width=1.5in] {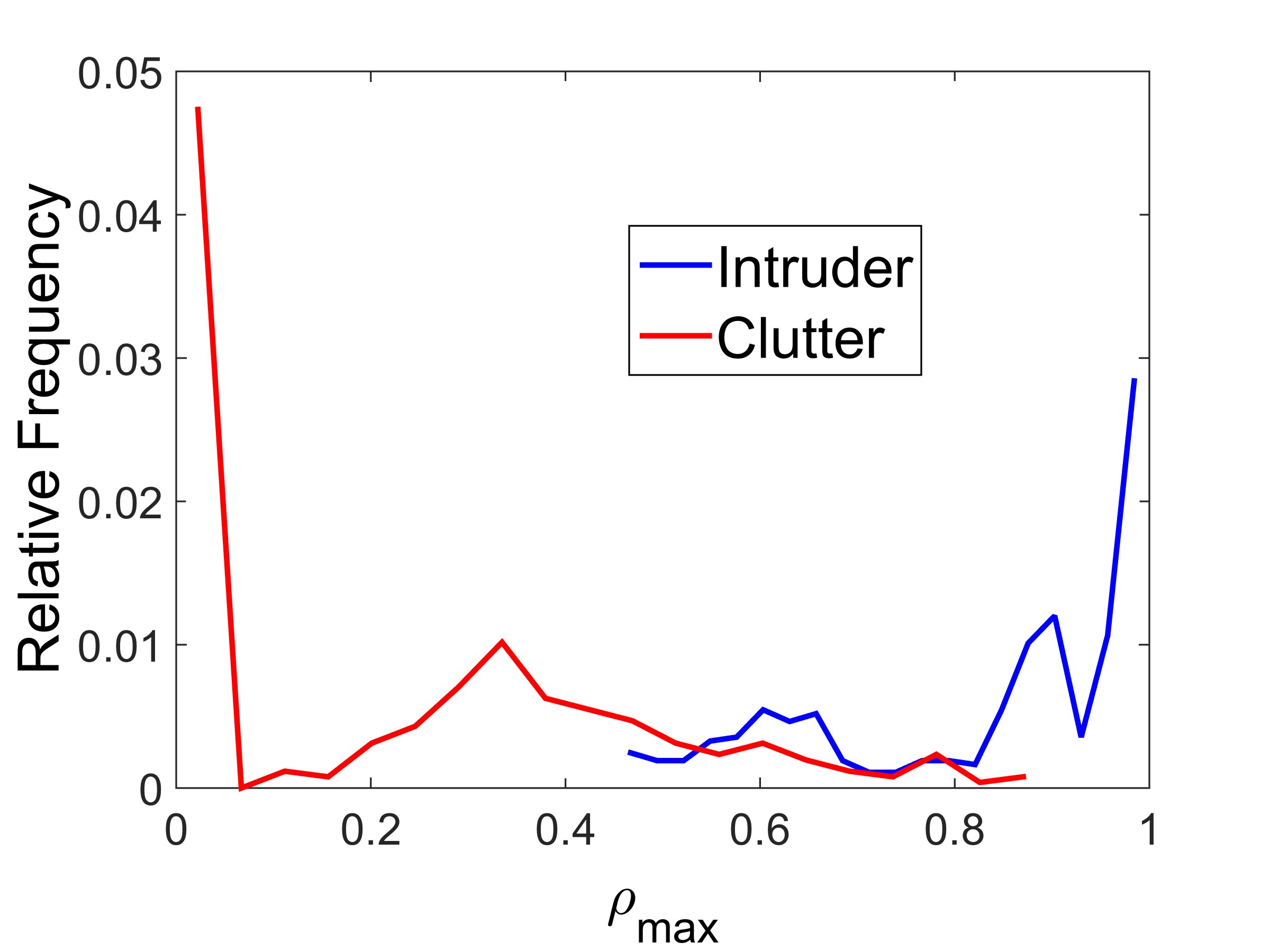}}
 \caption{Correlation histograms for real-world and simulated data.} 
 \end{figure}

Histogram plots of $\rho_{\max}$ observed for a large collection of events obtained both through real-world data collection and simulation by ASPIRE are shown in Figs.~\ref{fig:Correlation_fig_Real_Data} and\ref{fig:Correlation_fig_Anim_Data}, respectively.    Note that in both data sets, $\rho_{\max}$ tends to be high for intruder motion and small in the case of  vegetative motion.   The differences between the two plots can be attributed in part, to the fact that the simulation data were generated using a smaller number of shrubs (to keep the complexity of simulation manageable) than were encountered in the real world.   
In addition, animal motion in the real world tended to take place at a slower speed (at larger distances) than could be reliably measured by the STP. 
  \subsection{Signal-Energy-Based Classification}
Much effort went into designing the STP to endow it with a good spatial resolution.  As can be seen from the data in Table~\ref{tab:TruthTable}, it is possible to obtain some indication of the nature of the intruder and the intruder's relative distance simply by examining the energy of the signals generated in sensors $A,B,C,D$.  
  \vspace{-0.1in} 
 \begin{center}
	\begin{table}[h]
		\begin{center}
			\caption{Inference Drawn Based on Combination of Sensors $A,B,C,D$ that are Triggered.} \label{tab:TruthTable}
			\begin{tabular}{@{} ||m{0.1cm}|m{0.1cm}|m{0.1cm}|m{0.05cm}||m{5.25cm}|| @{}} \hline \hline 
					\multicolumn{4}{||c||}{Sensor} &  \multirow{2}{*}{Signal is indicative of} \\ \cline{1-4}
							A & B & C & D & {}   \\ \hline \hline 
							0 & 0 & 0 & 1 & short animal at 5 m \\ \hline 
							0 & 0 & 1 & 0 & animal at 10 m \\ \hline 
							0 & 0 & 1 & 1 & tall animal at  5 m \\ \hline 
							0 & 1 & 1 & 0 & human at 10 m\\ \hline
							0 & 1 & 1 & 1 & short human at 5 m \\ \hline 
							1 & 1 & 1 & 1 & human at 5 m \\ \hline 
	\multicolumn{4}{||c||}{(all other } &  clutter or else, combination unlikely\\ 
	\multicolumn{4}{||c||}{combinations)} &  \\ \hline \hline								
			\end{tabular}
		\end{center}
	\end{table}
\end{center}

In the table, a $0$ or $1$ entry indicates that the energy received by the particular sensor lies below or above a certain preset threshold, respectively. For example, a human intruder will typically trigger signals in all four sensors $A,B,C,D$, whereas an animal intrusion will most often trigger signals in sensors $C,D$.   If only sensors $A,B$ register signals, this is very likely indicative of  clutter motion.   Other inferences that can be drawn, also appear in the table.  In terms of implementation however, the energy from all 8 sensors are fed as a feature vector to an SVM.    An important advantage is the relative ease of extracting energy levels and hence the use of such features in classification is attractive from the standpoint of implementation on a mote that has limited computational capability.

\section{Classification Algorithms and Experimental Results} \label{sec:results}
  \vspace{-0.06in} 
\subsection{Database Description}
Real-world and simulated databases were prepared. Each database covered a large number of events corresponding to either human or animal intrusion or else, clutter. Each event corresponded to a collection of $(8 \times 1024)$ samples from the $8$ sensors with samples spaced 50 msec apart and hence corresponding to a time duration of approximately 50 secs.  The break-up of the databases corresponding to the different event classes is shown in Table~\ref{tab:Database}.  For each database, we used k-fold cross validation with $5$ folds. \footnote{The results presented in our previous paper \cite{upadrashta2015animation} used the holdout method for cross-validation and a smaller range for the parameter $C$ compared to what is employed in this paper.  The parameter $C$ is used in the SVM to control the trade-off between margin and allowed training errors.}
%
  \vspace{-0.06in} 
\subsection{Classification Algorithms} 
In the three-way classification considered here, there were three feature vectors employed which we will refer to as: 
\bean
E_8 & \Leftrightarrow & \text{the energy recorded by all $8$ sensors }, \\
\rho_{\max} & \Leftrightarrow  & \text{the maximum correlation parameter}, \\
\text{C}_{60} & \Leftrightarrow  & \text{the collection of $60$ chirplet parameters}.
\eean

The dimension of the feature vector $C_{60}$ is given by 
\bean
\text{($5$ parameters per chirplet)} \times \text{($3$ chirplets per sensor signal)} \\
\times \text{  ($4$ sensor signals) }  \ = \  \text{total of $60$ parameters}.
\eean
 \begin{center}
	\begin{table}[h]
		\begin{center}
			\caption{Event Distribution within the Database.} \label{tab:Database}
			\begin{tabular}{||c|c|c|c||} \hline \hline 
			\multicolumn{4}{||c||}{Real-World Data} \\ \hline \hline 
					 Human & Animal & Clutter & Total \\ \hline
							209 & 151 & 134 & 494 \\ \hline \hline 
			\multicolumn{4}{||c||}{Simulated Data} \\ \hline \hline 
				 Human & Animal & Clutter & Total \\ \hline
							210 & 186 & 272 & 668 \\ \hline \hline 							
			\end{tabular}
		\end{center}
	\end{table}
\end{center}
In terms of machine learning algorithm, we uniformly relied upon SVM.  We will use SVM($E_8$) and SVM($E_8 \cup \rho_{\max}$) to denote SVM-based classification algorithms that employed $E_8$ and $\left(E_8 \cup \rho_{\max}\right)$ as their respective feature vectors, etc.   We employed a $2$-step classification algorithm.  In the first step, we distinguished between intruder and clutter and in the next step we distinguish between human and animal intruder. 

\subsubsection{Distinguishing Between Intruder and Clutter} 
\begin{table*}[t]
	\centering
				\caption{Intruder Versus Clutter - Accuracy with Real-World Data} \label{tab:int_clu_real_data_results} 
	\begin{tabular}{@{} ||m{0.4cm}|m{0.6cm}|m{0.5cm}||m{1cm}|m{1cm}|m{1cm}||m{1cm}|m{1cm}|m{1cm}|| @{}} \hline \hline 						
		\multicolumn{3}{||c||}{Features} & \multicolumn{3}{c||}{Minimum Accuracy $\%$} & \multicolumn{3}{c||}{Average Accuracy $\%$} \\ \hline 
		$E_8$ & $\rho_{\max}$ & $C_{60}$ & Clutter & Intruder & Total & Clutter & Intruder & Total \\ \hline \hline 
			\color{dark-green}\checkmark &    & &  96.3  & 94.2 & 94.8 & 96.3 & 97.4 & 97.1 \\ \hline 
			\color{dark-green}\checkmark & \color{dark-green}\checkmark &  & 96.3 & 95.6 & 95.8 & 97.8 & 97.4 & 97.5 \\ \hline 
			 &  & \color{dark-green}\color{dark-green}\checkmark & 100 & 98.6 & 99.0 & 98.5 & 99.7 & 99.4 \\ \hline \hline
			\end{tabular}

	\end{table*}	
	\begin{table*}[t]
	\centering
				\caption{Intruder Versus Clutter - Accuracy with Simulated Data} \label{tab:int_clu_sim_data_results} 
	\begin{tabular}{@{} ||m{0.4cm}|m{0.6cm}|m{0.5cm}|m{1cm}|m{1cm}|m{1cm}|m{1cm}|m{1cm}|m{1cm}|| @{}} \hline \hline 						
		\multicolumn{3}{||c|}{Features} & \multicolumn{3}{c|}{Minimum Accuracy $\%$} & \multicolumn{3}{c||}{Average Accuracy $\%$} \\ \hline
		$E_8$ & $\rho_{\max}$ & $C_{60}$ & Clutter & Intruder & Total & Clutter & Intruder & Total  \\ \hline 
		\color{dark-green}\checkmark &  & & 87.0 & 88.6 & 88.0 & 92.6 & 92.9 & 92.8 \\ \hline 
			\color{dark-green}\checkmark & \color{dark-green}\checkmark &  & 87.0 & 96.2 & 92.5 & 94.1 & 95.0 & 94.6 \\ \hline 
			&  & \color{dark-green}\checkmark & 98.2 & 97.5 & 97.8 & 99.3 & 99.2 & 99.3 \\ \hline \hline
			\end{tabular}

	\end{table*}

This turned out to be the more challenging aspect of classification.  Here, we tried out several possible choices of features (see Fig.~\ref{fig:chirplet_classifier}) and the corresponding classification results can be found tabulated in Tables~\ref{tab:int_clu_real_data_results} and \ref{tab:int_clu_sim_data_results}.  We present results that represent the minimum and average accuracies attained.  Our findings can be summarized as follows:    \vspace*{-0.1in}
\begin{enumerate}[label=(\alph*)]
\item {\bf Improvement due to correlation parameter $\rho_{\max}$: } The performance of SVM($E_8 \cup \rho_{\max}$) was noticeably better than that of SVM($E_8$) showing the parameter $\rho_{\max}$ to be relevant in this context.
\item {\bf Efficacy of Chirplet Parameter set $C_{60}$: } 
\bit
\item Interestingly, SVM($C_{60}$) significantly outperformed SVM($E_8 \cup \rho_{\max}$) despite the fact that the former only makes use of data from sensors $A,B,C,D$.
\item In the context of $C_{60}$ data, both $\rho_{\max}$ and $E_8$ turned out to improve performance only marginally.   Based upon these results, the decision was made to select the classifier based only on $C_{60}$ as the feature vector. 
\eit 
\item {\bf Animation as a useful tool for algorithm testing: } The relative performance of the various algorithms on real-world and simulated data was very similar, that suggests animation is a very useful tool in this setting. The  performance of SVM($E_8$) on simulated data was lower compared to the performance on real-world data as some of the clutter events generated by ASPIRE were similar to intruder events. Such events were not present in the real-world database.
\een
\begin{figure}[h]
\begin{center}
\includegraphics[trim= 0in  0.25in 0in 0in, width=3.in]{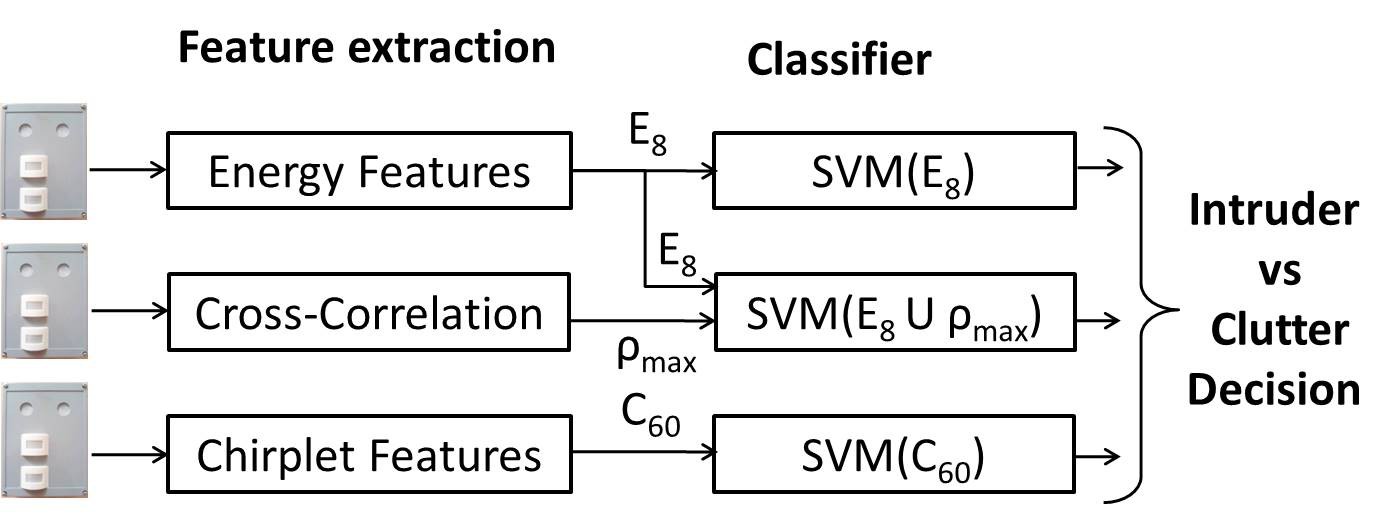}
\caption{Intruder vs clutter classification using energy, correlation and chirplet based feature vector.}
\label{fig:chirplet_classifier}
\end{center}
 \end{figure}
\subsubsection{Distinguishing Between Human and Animal}
In the second and final classification step, in order to distinguish between a human and animal, we simply fed the output of $E_8$ to an SVM that carried out the binary classification.  The spatial-resolution of the STP served us in good stead here, resulting in high-accuracy decisions.
\subsubsection{Classification Algorithm Finally Selected}
 Our 2-step classifier utilized the $C_{60}$ feature vector for intruder vs clutter classification. Upon detecting the presence of an intruder, $E_{8}$ features were then used to classify between human and animal in the second step (see Fig.~\ref{fig:final_classifier}). 
 \begin{figure}[h]
\begin{center}
\includegraphics[trim= 0in  0.25in 0in 0in, width=3.4in]{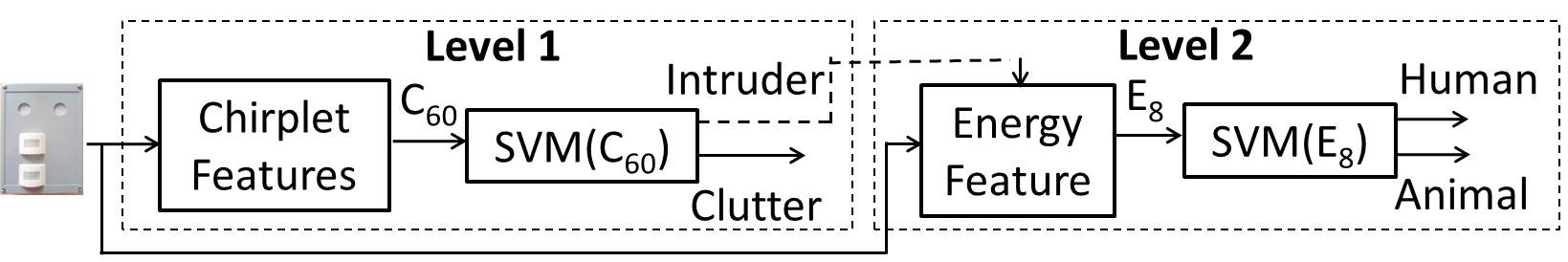}
\caption{Final 2-step classifier that utilizes $C_{60}$ and $E_{8}$ features.}
\label{fig:final_classifier}
\end{center}
 \end{figure} 

The minimum and average accuracies obtained on real-world and simulated data are shown in Table~\ref{tab:real_data_results_final}.
\begin{table}[h]
			\caption{Overall Classification Accuracy - 2-Step Classifier on Real-World and Simulated Data} \label{tab:real_data_results_final} 
		\begin{tabular}{@{} ||m{1cm}||m{1.2cm}|m{1.2cm}||m{1.2cm}|m{1.2cm}|| @{}} \hline \hline 	
			\multicolumn{1}{||c|}{} & \multicolumn{2}{c||}{ Real World Data } & \multicolumn{2}{c||}{Simulated Data } \\ \hline 
						
									& Minimum  & Average & Minimum  & Average \\
										& Accuracy &  Accuracy 	& Accuracy &  Accuracy\\
									 \hline 
									Clutter&  \text{ 92.3 \%}   & \text{ 97.8 \%}&  \text{ 98.1 \%}   & \text{ 99.6  \%} \\ \hline 
									Intruder&  \text{ 100 \%}   & \text{ 98.6 \%}&  \text{ 98.7 \%}   & \text{ 99.2  \%} \\ \hline 
									Human & \text{ 97.4 \%}  & \text{ 98.5 \%} &  100.0  \%  & 100.0  \% \\ \hline  
										Animal & 96.7  \%  & \text{ 98.6 \%} &  100.0  \%  & 100.0   \% \\ \hline \hline 
									Overall  & \text{ 95.8  \%}  & \text{ 97.3 \%} &  \text{ 98.5  \%}  & \text{ 99.4 \%} \\ \hline \hline
	\end{tabular}
	\end{table}
\section{Conclusions and Future Work}

The paper presents the design of a PIR STP that featured the use of (a) an array of PIR-sensors and lens combinations to endow the platform with the spatial resolution needed to distinguish between human, animal and vegetative clutter, (b) ASPIRE, a simulation tool that we developed that allowed us to judge the efficacy of various approaches to classification without the need for time consuming and challenging animal-motion data collection and (c) the modeling of an intruder signal as a waveform exhibiting chirp that was effective in discriminating between intruder motion and clutter.   The overall average classification accuracy was found to be quite high, over $97$\%.

The classification algorithm was carried out offline on a laptop as the complexity of extracting the $60$ chirplet parameters exceeds the computational capabilities of the processor on the mote.  It is planned as future work, to identify the key parameters among the $60$ chirplet parameters employed for classification here with the aim of significantly reducing the complexity of the classification algorithm to the point where it can be implemented on a mote.  
\vspace{-0.5cm}
\section{Acknowledgment}
We would like to acknowledge the help extended to us by the Executive Director and Staff at Bannerghatta Biological  Park and dog trainer at Dog Guru for animal data collection.
\vspace{-0.5cm}
\bibliographystyle{IEEEtran}
\bibliography{wsn_ITIT2}
\vspace{-1.2cm}	
\begin{IEEEbiographynophoto}{Raviteja Upadrashta} received the M.S. in Electrical Engineering from IIT Madras, Chennai in 2008. He is currently pursuing his Ph. D. in Department of Electrical and Communication Engineering at IISc, Bengaluru.
\end{IEEEbiographynophoto}
\vspace{-1.5cm}
\begin{IEEEbiographynophoto}{Tarun Choubisa} received the M.Tech. in Digital Signal Processing from IIT Guwahati in 2010. He is currently pursuing his Ph. D. in Department of Electrical and Communication Engineering at IISc, Bengaluru.
\end{IEEEbiographynophoto}
\vspace{-1.5cm}
\begin{IEEEbiographynophoto}{A. Praneeth(M'09)} received B.E. in Electronics and Communication from PES College of Engineering in 2003 and M. E. (Telecom) from IISc in 2015. He is currently working with DRDO.
\end{IEEEbiographynophoto}
\vspace{-1.5cm}
\begin{IEEEbiographynophoto}{Tony G.} received B. Tech. in Electrical and Electronics from NIT Calicut in 2013 and M. E. (Signal Processing) from IISc Bangalore in 2015. He is currently working in Flytxt Trivandrum as R \& D Lead Data Science.
\end{IEEEbiographynophoto}
\vspace{-1.5cm}
\begin{IEEEbiographynophoto}{V. S. Aswath} received B. Tech. from department of electrical engineering from NIT Calicut in 2011, and M. E. degree in Signal processing from IISc in 2013. Currently employed with Broadcom India Research. 
\end{IEEEbiographynophoto}
\vspace{-1.5cm}	
\begin{IEEEbiographynophoto}{P.    Vijay    Kumar
(S'80-M'82-SM'01-F'02) }
  received his  Ph.D.  from  USC  in  1983 in  Electrical  Engg.  From
1983 to  2003  he was  on the faculty  of the EE-Systems Department  at USC.
Since 2003, he has been on the faculty of IISc, Bengaluru. He also holds the position of
Adjunct  Research  Professor at USC.
His  current  research  interests  include  codes  for  distributed  storage  and
intrusion-detection  algorithms  for  WSNs.  He  is  an  ISI
highly  cited  author  and  a  Fellow  of  the
Indian  National  Academy  of  Engg. He is also co-recipient of the
1995 IEEE Information Theory Society
Prize-Paper  award,  a  Best-Paper  award  at  the  DCOSS  2008  conference  on
sensor networks and the IEEE Data Storage Best-Paper Award of 2011/2012.
A  pseudo-random  sequence  family  designed  in  a  1996  paper  co-authored
by  him  now  forms  the  short  scrambling  code  of  the  3G  WCDMA  cellular
standard.  He  received  the  USC  School  of  Engineering  Senior  Research
Award  in  1994  and  the  Rustum  Choksi  Award  for  Excellence  in  Research
in  Engg.  in  2013  at  IISc.  He  has  been  on  the  Board  of  Governors  of
the  IEEE  Information  Theory  Society  since  2013.
\end{IEEEbiographynophoto}
\vspace{-1.5cm}
\begin{IEEEbiographynophoto}{Sripad Kowshik} received B.E. in Electronics and Communication from Sri Venkateshwara College of Engg. in 2012. He is currently pursuing M.S. in Electrical Engg. and Computer  Science at University of California, Irvine. 
\end{IEEEbiographynophoto}
\vspace{-1.5cm}
\begin{IEEEbiographynophoto}{Hari Prasad Gokul R} received the B.E. degree in Electronics and Communications Engineering from PSG College of Technology (Anna University), Coimbatore, in 2013 and is currently a Project Assistant with the Department of Electronic Systems Engineering, IISc, Bangalore, India.
\end{IEEEbiographynophoto}
\vspace{-1.5cm}
\begin{IEEEbiographynophoto}{Prabhakar Venkata}
received M.Sc. (Engg.) from IISc, Bengaluru and PhD from TUDelft, Netherlands. He works as Senior Scientific Officer in the Department of Electronic Systems Engg, IISc, Bengaluru.  
His area of work is in Networked Embedded Systems.  His research interest is in Energy Harvesting systems, Power Management Algorithms, Tactile IoT. The broad spectrum
comprises of Modelling, Virtual Prototyping, System Building and Performance evaluation.  His work in LED based communication won the best demo award in COMSNETS 2014.  He is currently working on RFID localization algorithms, RF energy harvesting technologies in chip design and Indoor localization applications in healthcare and safety. The Zero Energy Networks laboratory (ZENLab) at IISc
specializes in building ultra low power embedded boards and software stacks.  The application areas are related to Smart Grids, Healthcare, Human Security and Agriculture.
\end{IEEEbiographynophoto}

\end{document}